\documentclass[lettersize,journal]{IEEEtran}
\usepackage[dvipsnames]{xcolor}
\usepackage[switch]{lineno} 

\usepackage{cite}
\usepackage{amsmath,amssymb,amsfonts}
\usepackage{algorithmic}
\usepackage{algorithm}
\usepackage{graphicx}
\usepackage[caption=false,font=footnotesize]{subfig}
\usepackage{textcomp}
\usepackage[normalem]{ulem}
\usepackage{xcolor}
\usepackage{bm}
\usepackage[flushleft]{threeparttable}
\usepackage{multirow}
\usepackage{algorithmic}
\usepackage{booktabs}   
\usepackage{bbm}
\usepackage{enumitem}
\usepackage{mathtools}
\usepackage{hyperref}   
\newcommand*{\eg}{\emph{e.g.}{}}
\newcommand*{\ie}{\emph{i.e.}{}}

\renewcommand{\d}{\mathrm{d}}
\usepackage{url}

\def\BibTeX{{\rm B\kern-.05em{\sc i\kern-.025em b}\kern-.08em
    T\kern-.1667em\lower.7ex\hbox{E}\kern-.125emX}}

\makeatletter
\newcommand{\linebreakand}{%
  \end{@IEEEauthorhalign}
  \hfill\mbox{}\par
  \mbox{}\hfill\begin{@IEEEauthorhalign}
}
\makeatother

\begin{document}

\title{
Global-Decision-Focused Neural ODEs for\\Proactive Grid Resilience Management
}
\author{Shuyi Chen,~\IEEEmembership{Student Member,~IEEE,} Ferdinando Fioretto, Feng Qiu,~\IEEEmembership{Senior Member,~IEEE,} Shixiang Zhu~\IEEEmembership{Member,~IEEE,}
}


\markboth{}%
{Shell \MakeLowercase{\textit{et al.}}: A Sample Article Using IEEEtran.cls for IEEE Journals}


\maketitle

\begin{abstract}
Extreme hazard events such as wildfires and hurricanes increasingly threaten power systems, causing widespread outages and disrupting critical services.
To mitigate these risks, grid operators must act proactively---pre-positioning crews and resources or scheduling hardening activities---guided by forecasts that directly support operational objectives.
Traditionally, utilities and agencies have followed a predict-then-optimize paradigm: first generating impact forecasts, then using them to inform response planning.
However, this two-step approach inherently separates forecasting from decision-making, overlooking how forecast errors propagate into downstream actions and often producing misaligned or suboptimal plans.
We address this gap with predict-all-then-optimize-globally (PATOG), a framework that unifies outage prediction with globally optimized interventions. At its core, our global-decision-focused (\texttt{GDF}) Neural ODE model jointly captures outage dynamics and optimizes pre-event resilience strategies in a decision-aware manner. Unlike conventional methods, our approach ensures spatially and temporally coherent decisions, enhancing both predictive accuracy and operational efficiency.
Evaluations on synthetic and real-world data show that \texttt{GDF} reduces decision regret by up to 75\% compared to two-stage baselines, enabling more effective proactive planning.
\end{abstract}


\section{Introduction}

Extreme hazard events such as wildfires, winter storms, hurricanes, and earthquakes can trigger widespread power outages, disrupting economic activity, threatening public safety, and complicating the delivery of critical services \cite{Handmer2012, Kenward2014}. 
For instance, the January 2025 wildfires in Los Angeles County, fueled by strong Santa Ana winds, led to the destruction of critical power infrastructure across vast areas, resulting in prolonged outages for over 400,000 customers \cite{reuters2024hurricane}. 
Similarly, Hurricane Milton in October 2024 made landfall in Florida, bringing extreme winds and flooding that damaged the electrical grid and left more than 3 million homes and businesses without power for weeks \cite{reuters2025wildfires}, illustrating the catastrophic consequences of extreme weather on energy resilience.

\begin{figure}[!t]
        \centering
        \includegraphics[width=\columnwidth]{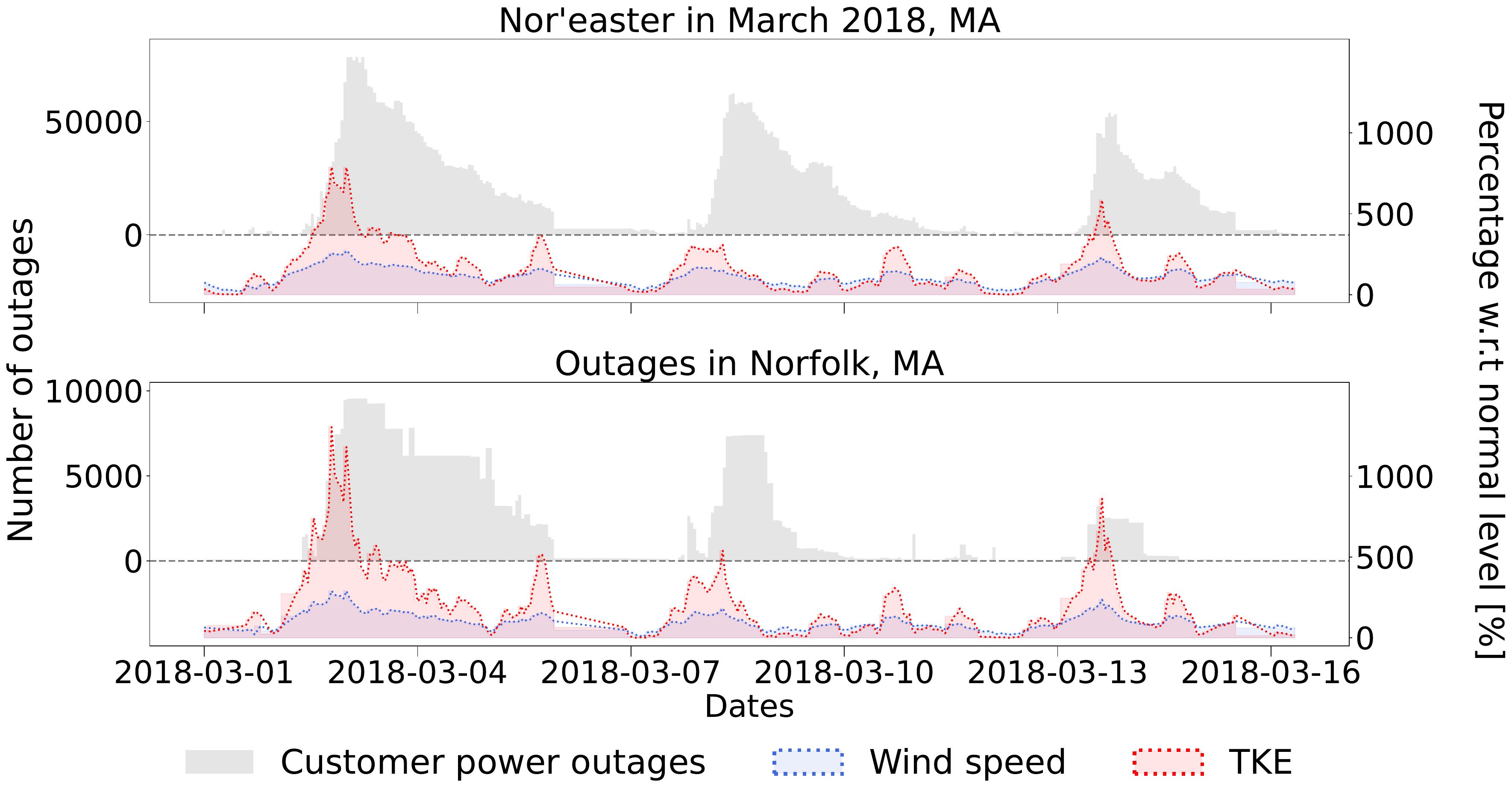}
        \caption{The number of outaged customers and the meteorological factors, including wind speed and turbulent kinetic energy.
        }
        \label{fig:vis}
\end{figure}

A fundamental challenge in strengthening power system resilience lies in the immense difficulty of restoring electricity after a severe event has caused widespread damage. The recent Los Angeles wildfires highlight how extreme winds can overwhelm firefighting efforts, making containment impossible even with abundant resources, let alone the rapidly repairing damaged transmission lines \cite{CNN2025Wildfires}.
The aftermath of major storms presents similar challenges---damaged infrastructure can take weeks or even months to restore, leading to severe economic and social repercussions. However, the impact of such events can be significantly mitigated, and in some cases, entirely avoided through proactive resilience planning. For example, preemptive de-energization strategies have been successfully implemented in California to reduce wildfire risks \cite{wiki:2019_CA_power_shutoffs}, while grid hardening and strategic upgrades have improved system resilience in hurricane-prone regions \cite{HoustonChronicle2025Grid}.

To proactively mitigate power grid disruptions, it is crucial to anticipate the long-term impact of extreme events on power systems and implement preemptive measures accordingly. Accurate forecasting enables utility companies and policymakers to assess risks, prioritize interventions, optimize grid reinforcement, and allocate resources efficiently. By identifying vulnerable regions in advance, decision makers can develop strategies that not only reduce power disruptions but also strengthen overall grid resilience.

A widely used approach in such planning is the predict-then-optimize (PTO) framework, where forecasts inform downstream decisions \cite{elmachtoub_smart_2022}.
However, conventional PTO approaches often fall short in high-stake decision making problems \cite{mandi_decision-focused_2024}. A high-quality forecast does not necessarily translate into effective decision-making, as small errors in predictive models can propagate into significant inefficiencies in downstream optimization tasks \cite{fernandez2022causal, mandi_decision-focused_2024}. This issue arises because conventional models are trained independently of decision-making objectives, meaning they may prioritize minimizing forecasting errors without considering how those errors impact critical decisions. As a result, even small deviations in predictions can lead to costly misallocations of resources, delayed responses, and suboptimal grid resilience strategies \cite{10855832,10711393}. Addressing this gap requires a more integrated approach that directly aligns predictive modeling with decision-making objectives, ensuring that forecasts are not only accurate but also actionable for proactive grid management.

This challenge becomes even more pronounced in system-wide decision-making, where resilience planning depends on aggregating independent predictions across multiple service units. 
Ensuring consistency across these independent predictions is inherently challenging, as errors in any individual unit’s forecast can propagate and compound, ultimately undermining the global decision quality. 
For instance, consider a utility company preparing for an approaching hurricane: if outage risks are underestimated in one region and overestimated in another, resources such as backup generators, repair crews, and grid reinforcements may be misallocated. Low-risk areas may be overprepared, while high-risk areas may lack sufficient resources, worsening service disruptions. 
These inconsistencies underscore the critical need for a holistic, system-wide approach that integrates predictions for multiple units with a global decision-making objective, ensuring coherent and globally optimized resilience strategies rather than fragmented, unit-specific responses. 

To address the challenges of proactive resilience planning, we first propose a new decision-making paradigm, referred to as predict-all-then-optimize-globally (PATOG), which integrates predictive outage modeling across all service units with a global decision-making process for grid resilience enhancement. 
Examples of such planning problems for better pre-event preparation include scheduling dispatch of mobile generators and planning budget-constrained power line undergrounding.
To solve PATOG problems, we introduce a global-decision-focused (\texttt{GDF}) neural ordinary differential equation (ODE) model, which simultaneously predicts the long-term evolution of system functionality (quantified by the number of power outages) and optimizes global resilience decisions. Inspired by epidemiological models, our method conceptualizes power outage progression within each service unit as a dynamical system, where the failure and restoration processes evolve based on transmission line conditions, local weather, and socioeconomic factors. These transition dynamics are parameterized using neural networks, allowing the model to adapt to different hazard scenarios and geographic conditions. 
Instead of treating each service unit independently, our method holistically forecasts the outage evolution across the power grid and derives a globally coordinated intervention strategy. This enables utilities to allocate resources more effectively, mitigating service disruptions in a spatially and temporally coherent manner.

To assess our framework, we conduct experiments on both real-world and synthetic datasets. The real dataset captures outage events from a 2018 Nor’easter in Massachusetts, combining county-level outage reports \cite{maema2020power} with meteorological data from NOAA’s HRRR model \cite{noaa_hrrr} and socioeconomic indicators from the U.S. Census Bureau \cite{USCensusBureau2017ACS} (Fig.~\ref{fig:vis}). The synthetic dataset simulates real outage propagation using a simplified SIR model, enabling controlled evaluation under varying weather conditions and grid configurations. These datasets provide a robust testbed for assessing \texttt{GDF} in realistic resilience planning scenarios.
Our results demonstrate significant improvements in forecast consistency, decision efficiency, and overall grid resilience, showcasing the potential of our approach to enable smarter, more proactive energy infrastructure management in the face of increasing hazard-induced disruptions.

We summarize our main contributions as follows: 
\begin{itemize}[left=0pt,parsep=0pt,itemsep=0pt]
    \item Propose a new predict-all-then-optimize-globally (PATOG) paradigm for system-wide decision-making problems in grid resilience management;
    \item Develop a novel method, referred to as global-decision-focused (\texttt{GDF}) Neural ODEs, for solving PATOG problems;
    \item Demonstrate that \texttt{GDF} Neural ODEs outperform baseline methods across multiple grid resilience management tasks through experiments on synthetic data.
    \item Evaluate our approach on a unique real-world customer-level outage dataset, uncovering key insights that inform more effective grid resilience strategies.
\end{itemize}


\subsection{Related Work}
\label{sec:ref}
This section reviews key advancements in outage modeling, decision-focused learning (DFL), and grid operation optimization, with an emphasis on their practical applications in power system resilience. Despite significant progress, existing methods separate forecasting from optimization, causing inefficiencies in decision-making. This review underscores the need for a unified framework that aligns predictive models with  resilience objectives to enhance grid reliability in extreme natural hazards.

\vspace{.05in}
\noindent\emph{Grid Operation Optimization}.
Grid optimization has been widely studied to improve power system resilience, covering areas such as distributed generator (DG) placement \cite{8668633,QIN2024752,9117454}, infrastructure reinforcement \cite{8273985}, and dynamic power scheduling \cite{5963580,990600}. However, traditional approaches often follow a two-stage predict-and-mitigate paradigm—first forecasting system conditions and then optimizing responses \cite{7080837}. This disconnect between prediction and optimization results in suboptimal grid operations, particularly under high uncertainty, where even small forecasting errors can cause significant deviations from the optimal response \cite{10711393,10855832}.


To overcome these limitations, we propose integrating decision-focused learning (DFL) with predictive modeling. By embedding decision objectives directly into the learning process, our approach aligns predictions with optimization goals, enabling more adaptive and proactive resource allocation and grid reinforcement. This integration enhances grid resilience amid escalating natural hazard risks.

\vspace{.05in}
\noindent\emph{Power Outage Modeling.}
Accurate power outage forecasting is crucial for enhancing grid reliability and resilience \cite{9160513, 7752978}. Various machine learning and statistical methods have been employed to predict outages under different conditions. 
These approaches incorporate neural networks enriched with environmental factors and semantic analysis of field reports, providing real-time updates and enhancing predictive performance through text analysis.

Additionally, ordinary differential equations have been widely used to model dynamic systems, such as outage propagation, capturing evolving disruptions under various conditions \cite{zhang2024recurrent,chen2018neural}. For instance, adaptations of the Susceptible-Infected-Recovered (SIR) model from epidemiology have been applied to simulate outage propagation, drawing parallels between power failures and disease spread \cite{9646114}.

While these models provide valuable insights, they often lack the granularity needed for city- or county-level decision-making, limiting their practical application to localized resilience planning. By integrating local weather forecasts and socio-economic data into compartmental Neural ODE models, our approach offers forecasting of local outage dynamics, enabling more targeted and effective interventions.

\vspace{.05in}
\noindent\emph{Decision-Focused Learning}.
Decision-focused learning (DFL) integrates predictive machine learning models with optimization, aligning training objectives with decision-making rather than purely maximizing predictive accuracy. Unlike traditional two-stage approaches, where predictions are first generated and then used as inputs for optimization, DFL enables end-to-end learning by backpropagating gradients through the optimization process. This is achieved via implicit differentiation of optimality conditions such as KKT constraints \cite{amos2017optnet, gould2021deep} or fixed-point methods \cite{kotary2023folded, wilder2019end}. For nondifferentiable optimizations, approximation techniques such as surrogate loss functions \cite{elmachtoub_smart_2022}, finite differences \cite{vlastelica2019differentiation}, and noise perturbations \cite{berthet2020learning} have been developed. Recent work has also explored integrating differential equation constraints directly into optimization models, enabling end-to-end gradient-based learning while ensuring compliance with system dynamic constraints \cite{Fioretto:ICLR25, jacquillat2024branch}.

A well-studied class of DFL problems involves linear programs (LPs), where the Smart Predict-and-Optimize (SPO) framework \cite{elmachtoub_smart_2022} introduced a convex upper bound for gradient estimation, enabling cost-sensitive learning for optimization. Subsequent work has extended DFL to combinatorial settings, including mixed-integer programs (MIPs), using LP relaxations \cite{mandi2020smart, wilder2019melding}. 
Recent advances, such as decision-focused generative learning (Gen-DFL) \cite{wang2025gendfldecisionfocusedgenerativelearning}, tackle the challenge of applying DFL in high-dimensional setting by using generative models to adaptively model uncertainty.

\vspace{.05in}
\noindent\emph{Differentiable Optimization}.
A key enabler of DFL is differentiable optimization (DO), which facilitates gradient propagation through differentiable optimization problems, aligning predictive models with decision-making objectives \cite{cvx}. Recent advances extend DO to distributionally robust optimization (DRO) for handling uncertainty in worst-case scenarios, improving decision quality under data scarcity \cite{zhu2022distributionally, chen2025uncertainty}. Beyond predictive modeling, DO has advanced combinatorial and nonlinear optimization through implicit differentiation of KKT conditions \cite{amos2017optnet}, fixed-point methods \cite{kotary2023folded}, and gradient approximations via noise perturbation \cite{berthet2020learning} and smoothing \cite{vlastelica2019differentiation}. These techniques bridge forecasting with optimization, ensuring decision-aware learning. In this work, DO enables the backpropagation of resilience strategy losses, aligning the spatio-temporal outage prediction model with grid optimization objectives.

\section{Predict All Then Optimize Globally}
\label{sec:patog}
In this section, we first provide an overview of decision-focused learning (DFL) and its application to solving predict-then-optimize (PTO) problems. We then introduce a new class of decision-making problems termed \emph{predict-all-then-optimize-globally} (PATOG). 
Unlike traditional PTO approaches, which generate instantaneous or overly aggregated forecasts and optimize decisions independently, PATOG explicitly accounts for how predictions evolve over time and space, integrating them into a single, system-wide optimization framework.

PATOG is particularly useful for grid resilience management, where decisions must consider complex interactions across all service units. A key example is the mobile generator deployment problem. In a conventional PTO setting, potential damage from an extreme weather event is first forecasted for each unit independently. Decisions, such as scheduling power generator deployments, are then made in isolation, without considering the evolving conditions of other units. This localized approach often leads to resource misallocation and suboptimal resilience outcomes.
In contrast, PATOG embeds these interdependencies into a global optimization problem, enabling system-wide decision-making that improves predictive models by incorporating cross-unit interactions. This results in more robust and effective resilience planning.

\subsection{Preliminaries: Decision-Focused Learning}

The predict-then-optimize (PTO) has been extensively studied across a wide range of applications \cite{7080837, mandi_decision-focused_2024}.  
It follows a two-step process: First, predicting the unknown parameters $\mathbf{c}$ using a model $f_\theta$ based on the input $\boldsymbol{z}$, denoted as $\hat{\boldsymbol{c}} \coloneqq f_\theta(\boldsymbol{z})$. Second, solving an optimization problem: 
\begin{equation}
    \boldsymbol{x}^*(\hat{\mathbf{c}}) = \arg \min_{\boldsymbol{x}} g(\boldsymbol{x}, \hat{\mathbf{c}}),
    \label{eq:pto}
\end{equation}
where $g$ is the objective function and $\boldsymbol{x}^*(\hat{\mathbf{c}})$ represents the optimal decision given the predicted parameters. 
This framework has numerous practical applications in power grid operations. 
For example, PTO is used in power grid operations to predict system stress using synchrophasor data, optimize outage management by forecasting disruptions and proactively dispatching restoration crews, and enhance renewable integration by predicting fluctuations in wind and solar generation to improve scheduling and grid balancing \cite{7080837}.
These predictive insights enable system operators to make informed strategic decisions, enhancing grid reliability and resilience.

However, the conventional two-stage approach does not always lead to high-quality decisions. 
In this approach, the model parameter $\theta$ is first trained to minimize a predictive loss, such as mean squared error. Then the predicted parameters $\hat{\mathbf{c}}$ are used to solve the downstream optimization. 
This separation between prediction and optimization can result in suboptimal decisions, as the prediction model is not directly optimized for decision quality \cite{kotary2021end}. 

To address this limitation, the decision-focused learning (DFL) integrates prediction with the downstream optimization process \cite{mandi_decision-focused_2024}.
Instead of optimizing for predictive accuracy, DFL trains the model parameter $\theta$ by directly minimizing the \emph{decision regret} \cite{mandi_decision-focused_2024}:
\[
\theta^* = \arg \min_{\theta} \mathbb{E} \left[ g(\boldsymbol{x}^*(f_\theta(\boldsymbol{z})), \mathbf{c}) - g(\boldsymbol{x}^*(\mathbf{c}), \mathbf{c}) \right].
\]
This approach ensures that the model is learned with the ultimate goal of improving decision quality, making it particularly effective for PTO problems.
Note that we assume that the constraints on decision variable $\boldsymbol{x}$ are fully known and do not depend on the uncertain parameters $\mathbf{c}$ in this study. This assumption simplifies the problem by ensuring that all feasible solutions $\boldsymbol{x}$ remain valid regardless of the parameter estimates.

\subsection{Proposed PATOG Framework}
\label{sec:proposed_patog}

The objective of PATOG in this work is to develop proactive global recourse actions that enable system operators to better prepare for natural hazards. 
These actions may include preemptive dispatch of mobile generators, strategic load shedding, grid reconfiguration, or reinforcement of critical infrastructure.
The PATOG consists of two steps:
($i$) Predicting the temporal evolution of unit functionality across all the service units in the network throughout the duration of a hazard event. 
($ii$) Deriving system-wide strategies that minimize overall loss based on all the predictions, enabling optimized resource allocation by anticipating critical failures before they occur.

Consider a power network consisting of $K$ geographical units, where each unit $k$ serves $N_k$ customers.
We define the global recourse actions as $\boldsymbol{x} \coloneqq \{x_k\}_{k=1}^K$, where $x_k$ represents the action taken for unit $k$. 
A key challenge in designing effective actions is understanding how the system will respond to  an impending hazard event. 
To this end, we use the number of customer power outages, which is publicly accessible via utility websites, as a measure of system functionality \cite{7080837}. 

To model the outage dynamics, we represent the outage state of each unit $k$ using a dynamical system over the time horizon $[0, T]$ during a hazard event.
During the event, the state of each unit $k$ at time $t$ is represented by three quantities: 
\begin{itemize}[left=0pt,itemsep=0pt]
    \item $U_k(t) \in \mathbb{Z}_*$: the number of unaffected customers;
    \item $Y_k(t) \in \mathbb{Z}_*$: the number of customers experiencing outages;
    \item $R_k(t) \in \mathbb{Z}_*$: the number of restored customers.
\end{itemize}
The total number of customers, $N_k$, in the unit $k$ remains constant throughout the event, satisfying the following constraint: 
\[
    U_k(t) + Y_k(t) + R_k(t) = N_k, ~\forall t \in [0, T].
\]
For compact representation, we define the state vector for unit $k$ as $\mathbf{S}_k(t) \coloneqq [U_k(t), R_k(t), Y_k(t)]^\top$. 
To simplify notation, we collectively represent the outage states across all units over the time horizon as $\mathbf{S} \coloneqq \{\mathbf{S}_k(t) \mid t \in [0, T] \}_{k=1}^K$.

Formally, we are tasked to search for the optimal action:
\begin{align}
    \boldsymbol{x}^*(\mathbf{S}) = &~\arg \min_{\boldsymbol{x}}~g(\boldsymbol{x}, \mathbf{S}) \label{eq:opt}\\
    \text{s.t.} \quad \frac{\d \mathbf{S}_k(t)}{\d t} = &~ f_\theta(\mathbf{S}_k(t), \boldsymbol{z}_k),~\forall k, \label{eq:ode}\\
    \mathbf{S}_k(0) = &~ [N_k, 0, 0]^\top,~\forall k, \label{eq:init-ode}
\end{align}
where $g$ quantifies the decision loss of the action $\boldsymbol{x}$ based on the predicted future evolution of outage states $\mathbf{S}$. 
The transition function $f_\theta$ models the progression of unaffected, restored, and outaged customers in each unit over time, influenced by both local weather conditions and socioeconomic factors, jointly represented as covariates $\boldsymbol{z}_k \in \mathbb{R}^p$.
We note that most power outages during extreme weather stem from localized transmission line damage, with cascading failures being rare \cite{zhu2021quantifying,10479971}. Thus, we assume each unit evolves independently under its local conditions.

We emphasize that \eqref{eq:opt} extends the traditional PTO framework by integrating predictions across multiple units over an extended future horizon to derive a single, globally optimized solution. Unlike PTO, which focuses only on instantaneous or localized dynamics, PATOG captures both temporal and spatial outage states while modeling how decisions influence outage dynamics across the entire system. 
This comprehensive approach enables proactive, system-wide high-resolution resilience strategies that adapt to evolving conditions. 

\section{Global-Decision-Focused Neural ODEs}
\label{sec:gdf}
\begin{figure}[!t] \centering \includegraphics[width=\linewidth]{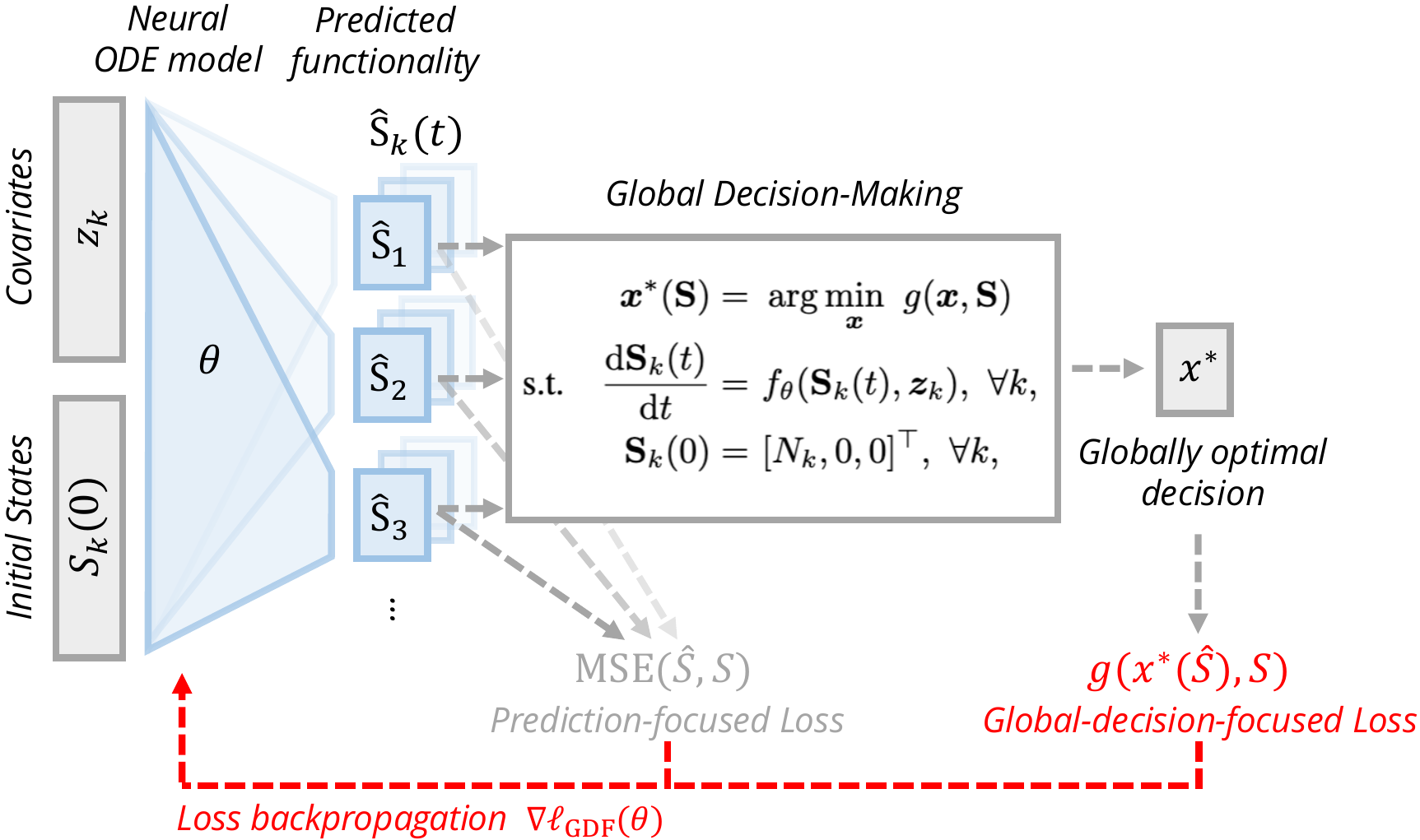} 
\caption{Overview of the proposed \texttt{GDF} framework. Given covariates $\boldsymbol{z}_k$ and initial states $\mathbf{S}_k(0)$ for all $K$ service units, a model parameterized by $\theta$ predicts the system states $\hat{\mathbf{S}}_k$ for all units. These predictions inform global decision-making, where optimal actions $\boldsymbol{x}^*(\hat{\mathbf{S}})$ minimize the global decision loss $g(\boldsymbol{x},\hat{\mathbf{S}})$. The framework optimizes $\theta$ by minimizing a global-decision-focused loss, regularized by a prediction-focused loss (\eg, MSE loss) to enhance predictive interpretability. Red arrows denote the backpropagation through $\nabla \ell_\texttt{GDF}(\theta)$, ensuring that the model learns both system-level decision quality and region-specific prediction accuracy. 
}
\label{fig:arc}
\end{figure}

This section presents a novel decision-focused neural ordinary differential equations (ODE) model tailored for solving PATOG problems in grid resilience management. The proposed Neural ODE model predicts outage progression at the unit level while simultaneously optimizing global operational decisions by learning model parameters in a decision-aware manner. We refer to this approach as global-decision-focused (\texttt{GDF}) Neural ODEs.
Fig.~\ref{fig:arc} provides an overview of the proposed framework.

\subsection{Neural ODEs for Power Outages}

Assume that we have observations of $I$ natural hazards (\eg, hurricanes, and winter storms) in the history. For the $i$-th event, the observations in unit $k$ are represented by a data tuple, denoted by $(\boldsymbol{z}_k^i, \{y_k^i(t)\})$, where $\boldsymbol{z}_k^i$ represents the covariates for unit $k$ during the $i$-th event and $y_k^i(t)$ is the number of customers experiencing power outages at time $t$. 
The outage trajectory $\{y_k^i(t) \mid t\in[0,T^i]\}$ is recorded at $15$-minute intervals.
A significant challenge in modeling outage dynamics is the lack of detailed observations for the underlying failure and restoration processes. 
Specifically, while $y_k^i(t)$ provides the number of customers experiencing outages at time $t$, we do not directly observe the failure and restoration states, $U_k(t)$ and $R_k(t)$, respectively. 

Drawing inspiration from the Susceptible Infectious Recovered (SIR) models commonly used in epidemic modeling \cite{kosma2023neural}, we conceptualize power outages within a unit as the spread of a ``virus''. 
In this analogy, outages propagate among customers due to local transmission line failures, while restorations provide lasting resistance to subsequent outages.
We formalize this analogy with the following three assumptions:
($i$) The number of unaffected customers $U_k(t)$ decreases over time as some customers transition from being unaffected to experiencing outages. 
This transition is governed by the \emph{failure transmission rate}, denoted by $\phi(\boldsymbol{z}_k; \theta_U)$.
Inspired by epidemiological transmission rates, this rate quantifies how local conditions---such as weather patterns and other regional factors encapsulated by covariates $\boldsymbol{z}_k$---influence the rate at which outages spread within the grid.
($ii$) Conversely, the number of customers with restored power ($R_k(t)$) gradually increases as the system operator repairs transmission lines and restores service. 
This process is captured by the \emph{restoration rate}, denoted $\phi(\boldsymbol{z}_k; \theta_R)$.
Both transmission and restoration rates, $\phi(\boldsymbol{z}_k; \theta_U)$ and $\phi(\boldsymbol{z}_k; \theta_R)$, are functions of the local covariates $\boldsymbol{z}_k$, and are modeled using deep neural networks.
($iii$) The total number of customers within each unit remains constant throughout the studied period. 
Based on these assumptions, we model the outage state transition for each unit $k$ in \eqref{eq:ode}, \ie, $\d \mathbf{S}_k(t) / \d t$, as follows:
\begin{equation}
\begin{cases}
\begin{aligned}
    \d U_k(t) / \d t &= -\phi(\boldsymbol{z}_k; \theta_U) Y_k(t) U_k(t), \\
    \d R_k(t) / \d t &= \phi(\boldsymbol{z}_k; \theta_R) Y_k(t), \\
    \d Y_k(t) / \d t &= -\d U_k(t) / \d t - \d R_k(t) / \d t.
\end{aligned}
\end{cases}
\label{eq:deterministic sys}
\end{equation}
These ODEs capture the dynamic evolution of power outage states within each unit $k$.
For notational simplicity, we use $\theta \coloneqq \{\theta_U, \theta_R\}$ to denote their parameters jointly.
This representation is particularly suitable for modeling weather-driven outage events with evolving spatio-temporal dynamics, such as those caused by winter storms or hurricanes, where the covariate-dependent rates $\phi(\boldsymbol{z}_k;\theta_U)$ and $\phi(\boldsymbol{z}_k;\theta_R)$ can capture exogenous meteorological forces leading to the hazard events.

In practice, we work with discrete observations $\{t_j\}_{j=0}^T$ and adopt the Euler method to approximate the solution to the ODE model \cite{chen2018neural}, \ie, 
\[
    \mathbf{S}_k(t_{j+1}) = \mathbf{S}_k(t_j) +  f_\theta(\mathbf{S}_k(t_j), \boldsymbol{z}_k) \Delta t_j,\quad\forall k,
\]
where $\mathbf{S}_k(t_{j+1})$ and $\mathbf{S}_k(t_{j})$ are the history states at times $t_{j+1}$ and $t_j$, respectively, 
and $\Delta t_j \coloneqq t_{j+1} - t_j$ is time interval.

\subsection{Global-Decision-Focused Learning}

The training loss for \texttt{GDF} Neural ODEs is formulated as the aggregated regret across all unit $k$, with an additional regularization term to ensure stability in prediction:
\begin{equation} 
\begin{aligned}
\ell_\texttt{GDF}(\theta) \coloneqq 
&~ \frac{1}{I} \sum_{i} \left [ g\Big(\boldsymbol{x}^*(\hat{\mathbf{S}}^i), {\mathbf{S}}^i \Big) -g\Big(\boldsymbol{x}^*({\mathbf{S}}^i), \mathbf{S}^i \Big) \right ] +\\
& ~ \lambda \cdot \frac{1}{IKT} \sum_{i,k,j} \left[ y_k^i(t_j) - \hat{Y}_k^i(t_j) \right]^2,
\end{aligned}
\label{eq:opt_dfl_ERM}
\end{equation}
where ${\hat{\mathbf{S}}^i}$ is the predicted outages states across all units for event $i$.

The objective function \eqref{eq:opt_dfl_ERM} consists of two key components:
($i$) Global-decision-focused loss (first term): This term evaluates the quality of the optimal action based on predictions, capturing the impact of prediction errors on operational decisions. The loss and its gradient are computed over all geographical units affected by the event, ensuring that learning is guided by system-wide decision quality. However, this loss alone does not provide direct insights into the structure of outage trajectories, which may limit the interpretability of the learned model.
($ii$) Prediction-focused loss (second term): To address this limitation, a prediction-based penalty is introduced to minimize discrepancies between observed outage trajectories $y^i_k(t)$ and predicted values $\hat{Y}^i_k(t)$. This term refines the model's ability to capture outage dynamics without explicitly observing the failure and restoration processes.
A user-specified hyperparameter $\lambda$ governs the trade-off between prediction accuracy and the regret associated with suboptimal operational decisions.

The most salient feature of the proposed \texttt{GDF} Neural ODEs method is its incorporation of both global and local perspectives. The global-decision-focused loss aggregates errors across all geographical regions and time steps, directly linking prediction quality to system-wide resilience measures and operational strategies (\eg, resource dispatch, outage management, and service restoration). Meanwhile, the prediction-focused component refines local accuracy by penalizing deviations at each service unit. 
By incorporating predictive regularization, the model empirically improves generalization to new events with unknown distributional shifts, mitigating overfitting, particularly given the limited availability of extreme event data.

\subsection{Model Estimation}

The learning of \texttt{GDF} Neural ODEs is carried out through stochastic gradient descent, where the gradient is calculated using a novel algorithm based on differentiable optimization techniques \cite{amos2017optnet,cvx, zhu2022distributionally}.
To enable differentiation through the $\arg \min$ operator in \eqref{eq:opt} embedded in the global-decision-focused loss, we relax the decision variable $\boldsymbol{x}$ from a potentially discrete space to a continuous space. 
At evaluation, we re-solve the original mixed-integer problems (\eg, Eq.~\eqref{eq:initial_stock_cities}--\eqref{eq:integer_constraint} or Supplementary Material Eq.~(1)) using Gurobi \cite{gurobi}. All reported cost, SAIDI, and regret values are computed using integer solutions, ensuring full operational validity.

For combinatorial optimization problems (\eg, power line undergroungding), the problem is reformulated as a differentiable quadratic program, and a small quadratic regularization term is added to ensure continuity and strong convexity \cite{wilder2019melding}.
Formally, we replace the original objective in \eqref{eq:opt} with the following:
\begin{equation} 
\min_{\boldsymbol{x}} g(\boldsymbol{x}, \mathbf{S}) + \rho\|\boldsymbol{x}\|_2^2,
\label{eq:regularized-LP}
\end{equation}
where $\rho > 0$ ensures differentiability. The full training procedure is summarized in Algorithm~\ref{alg:gdfl}. More implementation details can be found in the Supplementary Material.

To improve training efficiency, we avoid computing \texttt{GDF} gradients at every iteration, since evaluating the decision loss is computationally costly. Instead, we first pretrain the Neural ODE with the standard MSE loss, which quickly guides the model toward a good predictive initialization using inexpensive gradient computations. Starting from this pretrained state, we then apply a few finetuning iterations with the \texttt{GDF} loss, thereby reducing the number of expensive decision-loss evaluations while still aligning the model with global decision-focused objectives.
Empirically, our training time remains comparable to that of RNN and LSTM, underscoring both the tractability and scalability of the proposed approach. More details are provided in Section~H of the Supplementary Material.



\begin{algorithm}[!t]
\caption{Learning of \texttt{GDF} Neural ODEs}
\label{alg:gdfl}
\textbf{Input:} Data $\mathcal{D}=\{(\boldsymbol{z}_k^i, \{y_k^i(t)\})\}$, initial parameter $\theta_0$, initial states $\mathbf{S}^i(0)$, learning rate $\eta$, trade-off parameter $\lambda$, epochs $N$.\\
\textbf{Output:} $\theta^*$
\begin{algorithmic}[1]
\label{alg:alg1}
\FOR{epoch $n=1,\dots,N$}
    \FOR{$i = 1$ to $I$}
\STATE Initialize: $\hat{\mathbf{S}}^i(0) \leftarrow \mathbf{S}^i(0)$.
        \FOR{$j=1$ to $T_i$}
            \STATE $\hat{\mathbf{S}}^i(t_{j}) \leftarrow \hat{\mathbf{S}}^i(t_{j-1}) + f_\theta(\hat{\mathbf{S}}^i(t_{j-1}),\boldsymbol{z}_k^i)\Delta t_{j-1}$
        \ENDFOR
        \STATE $\ell_{\texttt{GDF}}^i(\theta) \leftarrow g\big(\boldsymbol{x}^*(\hat{\mathbf{S}}^i),\mathbf{S}^i\big)-g\big(\boldsymbol{x}^*(\mathbf{S}^i),\mathbf{S}^i\big)$
        \STATE $\theta \leftarrow \theta - \eta\, \nabla_\theta \ell_{\texttt{GDF}}^i(\theta)$.
    \ENDFOR
    \FOR{each mini-batch $\mathcal{B}\subset\mathcal{D}$}
        \STATE 
        $
        \ell_\text{Pred}(\theta) \leftarrow \frac{1}{|\mathcal{B}|KT}\sum_{(i,k,t)\in\mathcal{B}}\Big[y_k^i(t)-\hat{Y}_k^i(t)\Big]^2
        $
        \STATE $\theta \leftarrow \theta - \eta\, \lambda\,\nabla_\theta \ell_\text{Pred}(\theta)$.
    \ENDFOR
\ENDFOR
\RETURN $\theta$
\end{algorithmic}
\end{algorithm}

\section{Application: Mobile Generator Deployment}
\label{sec:mobile}

\begin{figure}[t!]
\centering
\includegraphics[width=\linewidth]{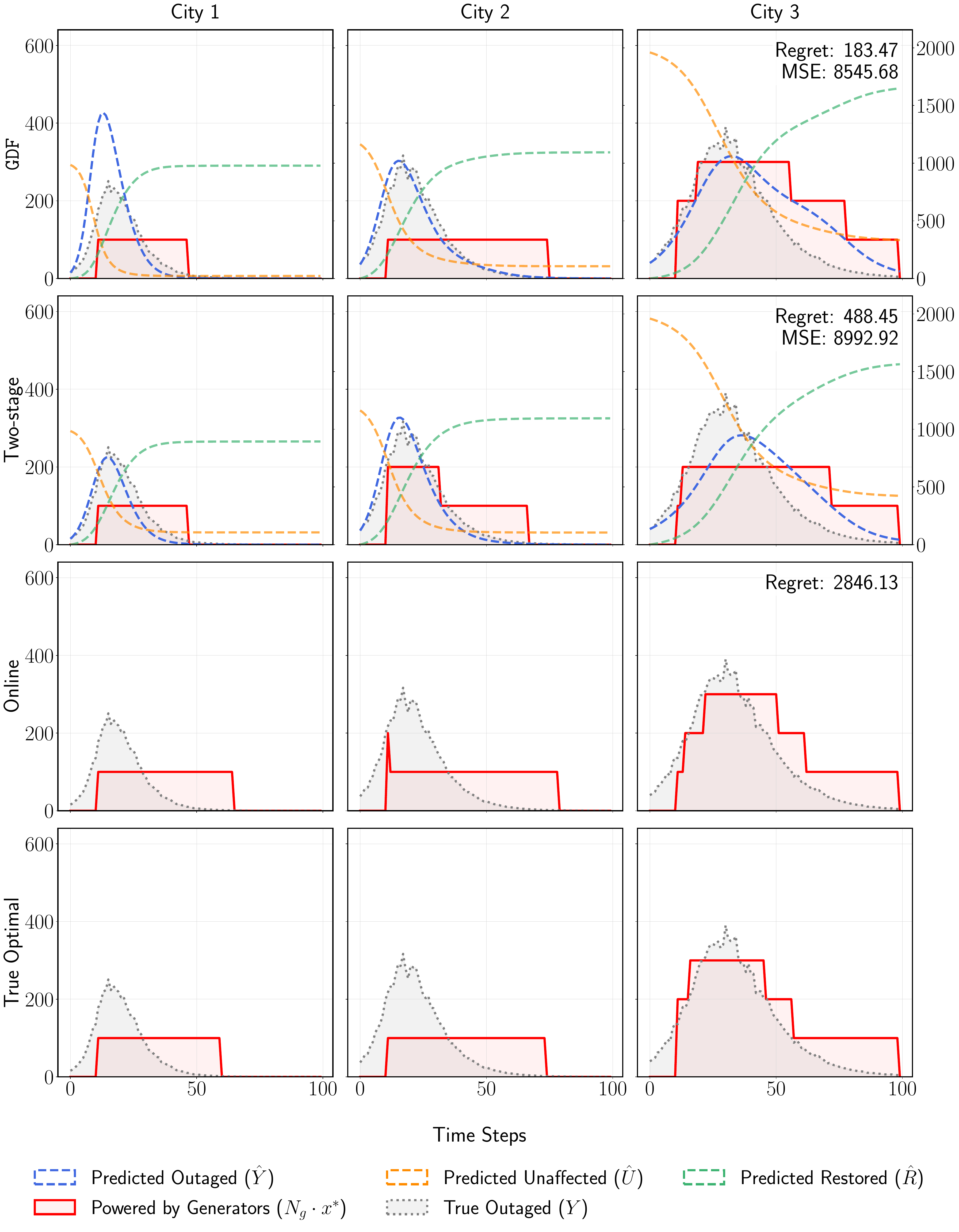}
\caption{A synthetic example of the mobile generator deployment problem for a system with three cities and five generators ($Q_w = 5$). The $y$-axis represents the number of outaged households. In this example, the uniform travel time $\delta_t = 10$, transportation cost is set to $c = 400$, the customer interruption cost is $\tau = 1$, and the operational cost is $\gamma = 2$. Four methods are compared on out‑of‑sample data: the proposed \texttt{GDF} framework (regret = $183.47$), a two‑stage approach (regret = $488.45$), an online baseline with observation lag of 5 (regret = $2846.13$), and the optimal solution with the ground-truth. Details of the synthetic data setup are provided in Section~\ref{sec:dataset}.}
\label{fig:diesel}
\end{figure}

Our method is broadly applicable to various proactive decision-making problems in grid operations, particularly in response to potential hazard events. 
This section highlights the adaptability of \texttt{GDF} by applying it to a mobile generator deployment task \cite{8668633,QIN2024752,9117454}, a representative \texttt{PATOG} problem. 

The objective of the mobile generator deployment problem is to strategically deploy mobile (\eg, diesel) generators across a network of potentially affected sites before a large-scale power outage to minimize associated costs. Due to the time-sensitive nature of power restoration, it is crucial to anticipate the spatiotemporal dynamics of outages while accounting for operational constraints such as generator capacity, fuel availability, and transportation logistics. \texttt{GDF} is well-suited for this task as it jointly learns outage patterns across cities over the planning horizon while optimizing deployment decisions, enabling proactive decision-making that adapts to evolving conditions.


Formally, let $Q_w \in \mathbb{Z}_+$ denote the initial inventory of generators at warehouse $w$.
For simplicity, we assume that warehouses also function as staging areas, where generators are initially stored and returned after deployment.
We assume that each generator has a fixed maximum capacity, capable of supplying electricity to $N_g$ customers, and that a single extreme event is anticipated.
The planning horizon is discretized into uniform time periods $\mathcal{T} = \{1, 2, \dots, T\}$, where ${T}$ represents the expected duration of the outage.
Let $\mathcal{K} = \{1, 2, \dots, K\}$ be the set of $K$ service units (\eg, cities or counties) where generators can be deployed, and $\mathcal{W} = \{1, \dots, W\}$ be the set of warehouses where all generators are initially stored. 
We denote the stock of generators in unit $k$ at time $t$ as $q_{tk}$. 

The mobile generator deployment problem is therefore defined on a directed graph denoted by $(\mathcal{V}, \mathcal{E})$, where $\mathcal{V} \coloneqq \mathcal{K} \cup \mathcal{W}$ is the vertex set, and $\mathcal{E}$ represents the edge set defining feasible transportation routes.
Each edge $(k, k') \in \mathcal{E}$ incurs a transportation cost $c_{kk'}$, and an associated travel time $\delta_{kk'} \in \mathbb{Z}_+$, representing the number of time periods required for generators to be transported from location $k$ to location $k'$.
The decision variables, ${\boldsymbol{x} = \{x_{tkk'}\}}$, $(k, k') \in \mathcal{E}$, $t \in \mathcal{T}$, specifies the transportation schedule, where $x_{tkk'} \in \mathbb{Z}_+$ represents the number of generators transported from location $k$ to $k'$ at time $t$. 
For simplicity, we assume negligible deployment time, allowing generators to become operational immediately upon arrival to supply power to $N_g$ customers.

The cost function of generator deployment problem is:
\begin{equation}
\begin{aligned}
    & g(\boldsymbol{x}, \mathbf{S}) \coloneqq \\
    & \underbrace{\sum_{(k,k') \in \mathcal{E}} \sum_{t=1}^T c_{kk'} x_{tkk'}}_{\text{Transportation Cost}} 
    + \underbrace{\gamma \sum_{k \in \mathcal{K}} \sum_{t=1}^T q_{tk} \vphantom{\sum_{(k,k') \in \mathcal{E}} \sum_{t=1}^T c_{kk'} x_{tkk'}}}_{\text{Operation Cost}}
    + \underbrace{h(\boldsymbol{x}, \mathbf{S}) 
    \vphantom{\sum_{(k,k') \in \mathcal{E}} \sum_{t=1}^T c_{kk'} x_{tkk'} + \gamma \sum_{k \in \mathcal{K}} \sum_{t=1}^T q_{tk}}}_{\text{Outage Cost}},
\end{aligned}
\label{eq:objective_Generator Distribution Problem}
\end{equation}
which consists of three components: 
($i$) Transportation Cost: The cost of delivering generators from warehouses to service units.
($ii$) Operation Cost: The fixed operational cost per unit time ($\gamma$) for deployed generators.
($iii$) Outage Cost: Economic loss due to power outages, determined by the number of customers experiencing outages given the generator deployment decision $\boldsymbol{x}$.
For simplicity, we define the outage cost as:
\[
    h(\boldsymbol{x}, \mathbf{S}) = \tau \sum_{k \in \mathcal{K}, t \in \mathcal{T}} \max \Big \{Y_k(t) - q_{tk} N_g, 0 \Big \},
\]
where $\tau$ denotes unit customer interruption cost, an adapted version of the Value of Lost Load (VoLL) \cite{6672826}, representing the economic impact per household outage per day, and $Y_k(t)$ denotes the number of outages in service unit $k$ at time $t$, extracted from the state $\mathbf{S}$ according to \eqref{eq:deterministic sys}.
This framework primarily evaluates system functionality based on the number of customers experiencing outages ($Y$). It is worth emphasizing that it is flexible to incorporate additional metrics, including the number of restored customers ($R$) and unaffected customers ($U$), if needed.

The transportation and inventory constraints of the mobile 
generator deployment are as follows:
\begin{align}
& q_{0k} = 0, \quad \forall k \in \mathcal{K} 
\label{eq:initial_stock_cities} \\
& q_{0w} = Q_w, \quad \forall w \in \mathcal{W} \label{eq:initial_stock_warehouse} \\
& q_{tk} = \sum_{k' \in \mathcal{V}} \sum_{\substack{t' \in \mathcal{T} \\ t' + \delta_{k'k} \leq t}} x_{t'k'k} - \sum_{k' \in \mathcal{V}} x_{tkk'}, \quad \forall k \in \mathcal{V}, \, t \in \mathcal{T}  \label{eq:stock_dynamics_warehouse} \\
& x_{tkk'} \leq C, \quad \forall t \in \mathcal{T}, (k,k') \in \mathcal{E} \label{eq:flow_between_cities} \\
& \sum_{k \in \mathcal{V}/{\{k'\}} , t \in \mathcal{T}} x_{tkk'} = \sum_{k \in \mathcal{V}/{\{k'\}}, t \in \mathcal{T}} x_{tk'k}, \quad \forall k' \in \mathcal{V}, \label{eq:flow_constraint}\\
& q_{kk'}, x_{tkk'} \in \mathbb{Z}_+, \, \quad  \forall (k,k') \in \mathcal{E}, \, \forall t \in \mathcal{T} \label{eq:integer_constraint}.
\end{align}
Constraints \eqref{eq:initial_stock_cities} and \eqref{eq:initial_stock_warehouse} establish the initial generator stock at each location.
Equation \eqref{eq:stock_dynamics_warehouse} tracks the generator stock at each location, accounting for the travel time $\delta$ taken for incoming shipments.
Inequality constraint \eqref{eq:flow_between_cities} limits the flow between service units or warehouses in $\mathcal{E}$ to a maximum capacity $C$.
Finally, equation \eqref{eq:flow_constraint} ensures flow conservation, requiring that the total inflow of generators equals the total outflow over the planning horizon for each node in the network. As a result, all generators return to the staging areas after the events.

To implement the deployment strategy based on our \texttt{GDF} framework, we first predict outage levels $\hat{Y}_k(t)$ for all service units using the Neural ODE model specified in \eqref{eq:deterministic sys}. 
The model is learned by minimizing the \texttt{GDF} loss defined in \eqref{eq:opt_dfl_ERM}. These global-deision-focused predictions are then integrated into the objective function \eqref{eq:objective_Generator Distribution Problem} of the mobile generator deployment problem, and the decisions are derived by solving a mixed-integer linear programming (MILP). 
Figure~\ref{fig:diesel} illustrates the optimal transportation schedules derived from the \texttt{GDF} predictions using an MILP solver, compared to the schedules produced by the Two-stage and reactive online approaches in a stylized example of the mobile generator deployment problem.



\section{Experiments}
\label{sec:exp}

In this section, we evaluate the proposed \texttt{GDF} on two grid resilience management problems—mobile generator deployment and power line undergrounding. The numerical results demonstrate its superior performance compared to conventional Two-stage methods, enabling better decision-making in the face of natural hazards.

\subsection{Dataset Overview}
\label{sec:dataset}
We evaluate \texttt{GDF} using both real and synthetic datasets to assess its effectiveness in outage prediction and resilience planning. The real dataset records the number of customers affected by outages during the 2018 Nor’easter in Massachusetts \cite{wiki:March_11-15_2018_nor'easter}, while the synthetic outage trajectories are generated using a simplified SIR model.
\subsubsection{Real Dataset}
The real dataset used in this study comprises county-level customer outage counts \cite{maema2020power}, combined with meteorological measurements and socioeconomic indicators from regions affected by a Nor’easter snowfall event in Massachusetts in 2018 (Fig.~\ref{fig:vis}). Meteorological variables—such as wind speed, temperature, and pressure—are sourced from NOAA’s High-Resolution Rapid Refresh (HRRR) model \cite{noaa_hrrr}.
Socioeconomic and demographic data are collected from the U.S. Census Bureau’s American Community Survey \cite{USCensusBureau2017ACS} and include median household income, median age, the number of food stamp recipients, the unemployment rate, the poverty rate, college enrollment, mean travel time to work, and average household size. These variables capture economic and mobility factors that may influence outage restoration dynamics and the effectiveness of emergency response efforts.

A key advantage of using outage data from Massachusetts during the 2018 Nor’easter event \cite{wiki:March_11-15_2018_nor'easter} is that three consecutive snowstorms impacted the power system within a short 15-day period, during which the local infrastructure remained largely unchanged. This allows us to reasonably assume that the outage patterns from these storms follow the same underlying data distribution, making them suitable for widely used train-test evaluation. Accordingly, we use the first storm for training and the second for testing the effects of different frameworks, while excluding the third storm due to its relatively minor impact. Fig.~\ref{fig:vis} illustrates the spatiotemporal dynamics of the real dataset.

\subsubsection{Synthetic Dataset}
To augment these real-world observations and enable experiments under varying conditions, synthetic outage trajectories are generated using a simplified SIR model. In this model, each county is treated as an independent population that experiences outages and eventually recovers, in accordance with the dynamics specified in \eqref{eq:deterministic sys}. Simulated weather conditions are incorporated to modulate the transmission rate, thereby capturing the variability and severity of extreme events. This synthetic dataset provides a realistic and flexible testbed for systematically evaluating the proposed decision-focused learning framework under different scenarios.

\subsection{Experimental Setup}
\label{sec:experiment_setup}

To evaluate the effectiveness of the proposed Global Decision-Focused (\texttt{GDF}) Neural ODE framework, we design our experiments in two stages:
($i$) testing the predictive accuracy and physical interpretability of the Neural ODE model itself, and
($ii$) applying it to two representative grid resilience planning and operations problems---power line undergrounding \cite{8278121, 7922545,AbiSamra2013,Shea2018} and mobile generator deployment \cite{8668633,QIN2024752,9117454}.
In both applications, the model produces predictions over the full planning horizon before the hazard occurs, reflecting the real-world challenge that utilities cannot rely on real-time data streams during rapidly evolving disasters \cite{9803820}. Once the hazard intensity exceeds a predefined threshold (set to 1\% in our experiments), the ODE model’s initial conditions are established, and the corresponding decisions are made.

The training of the Neural ODE follows a two-step process. First, the model is trained with a standard mean squared error (MSE) loss to establish a predictive baseline, from which optimal decisions are derived as a reference point. Building on this initialization, the model is then refined through end-to-end training using the combined loss function in \eqref{eq:opt_dfl_ERM}, which balances decision-focused and prediction-focused objectives, as detailed in Algorithm~\ref{alg:alg1}.

\begin{figure}[!t]
    \centering
    \includegraphics[width=\linewidth]{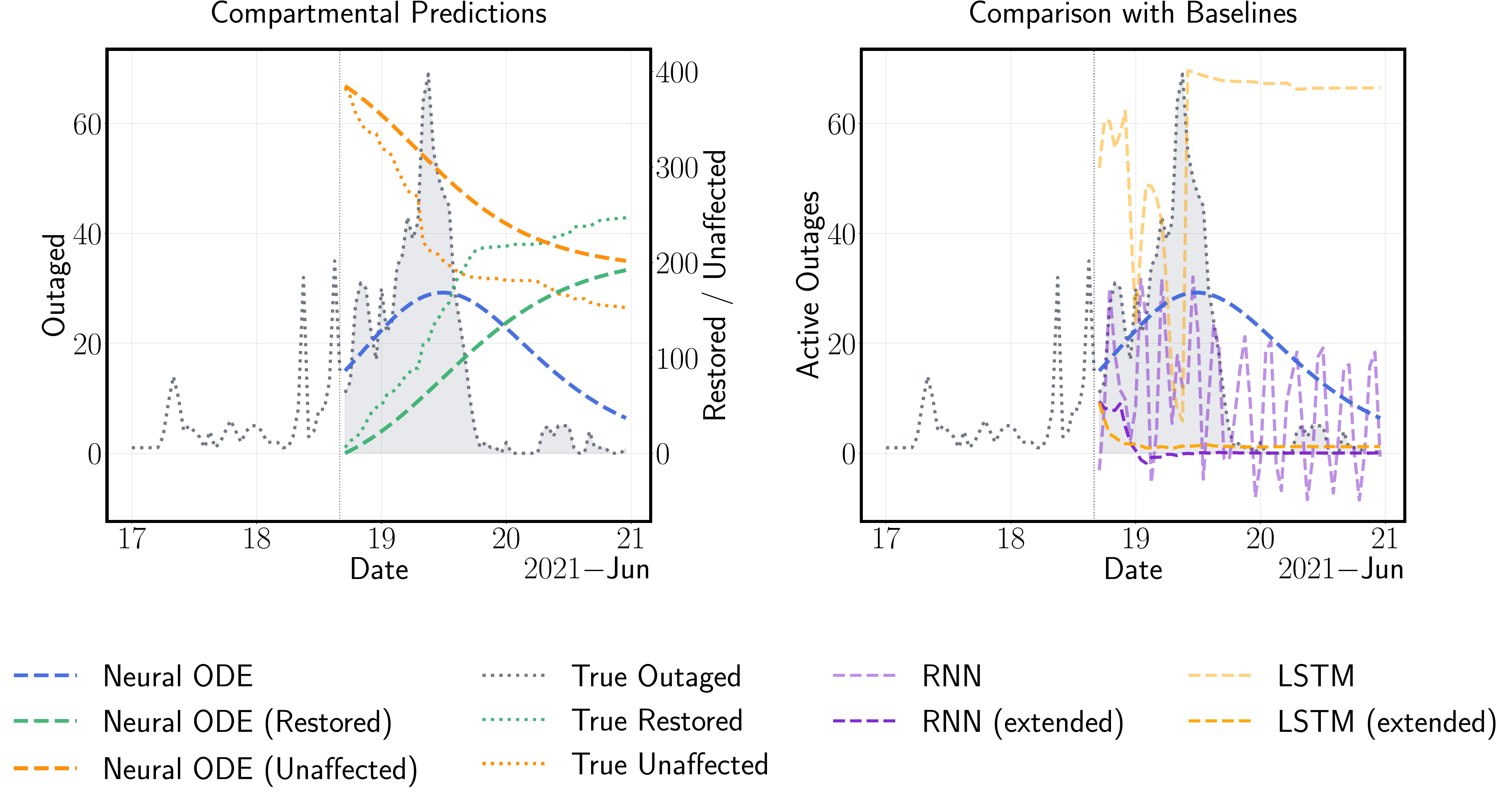}
    \caption{The real outage and restoration trajectories in Indianapolis, IN and their predictions. Left: actual outages and the Neural ODE prediction; the right axis shows restored and unaffected customers for both ground-truth and Neural ODE prediction. The vertical dotted line marks the train–test split. Right: comparison of test-set active-outage predictions from the Neural ODE, RNN (and extended), and LSTM (and extended).}
    \label{fig:indy-neuralode}
\end{figure}

We compare the proposed model, \texttt{GDF}, with spatio-temporal Neural ODE models trained solely on MSE loss (Two-stage) and the ground-truth optimal decision baseline (True Optimal). We also include results from an online method for the mobile generator deployment problem. Evaluation is based on three metrics: ($i$) prediction accuracy, measured as the mean squared error between predicted and observed outages; ($ii$) decision loss, represented by the total cost for generator deployment problem or the total System Average Interruption Duration Index (SAIDI) \cite{ieee1366_2022} for power line undergroudning problem; and ($iii$) regret, defined as the cost difference between the ground-truth optimal decision and the decision based on the model’s predictions. For all experiments on synthetic datasets, reported results and standard deviations are calculated with three different random seeds, which generate varying outage curves and weather conditions.

\begin{table}[!t]
\centering
\begin{threeparttable}
\caption{Outage prediction performance with Indianapolis Data}
\label{tab:added-experiments}
\renewcommand{\arraystretch}{1.15}
\begin{tabular}{@{}llll@{}}
\toprule
{Model} & {\# Params} & {MSE (Outage)} & {MSE (Restored)} \\
\midrule
\multicolumn{4}{l}{\textit{Short Period: 2021‑06‑17 -- 2021‑06‑20}} \\
RNN                     & 12{,}673 & 471.7          & N/A \\
RNN (extended)          & 12{,}673 & 577.2          & N/A \\
LSTM                    & 50{,}497 & 455.4          & N/A \\
LSTM (extended)         & 50{,}497 & 567.8          & N/A \\
Neural ODE              & \textbf{2}        & \textbf{252.7} & \textbf{3936.0} \\
\addlinespace
\multicolumn{4}{l}{\textit{Long Period: 2023-02-27 -- 2023-04-04}} \\
RNN                     & 12{,}673 & 1106.0         & N/A \\
RNN (extended)          & 12{,}673 & 1502.5         & N/A \\
LSTM                    & 50{,}497 & 327.6          & N/A \\
LSTM (extended)         & 50{,}497 & 225.6          & N/A \\
Neural ODE              & \textbf{1,986}    & \textbf{93.5}  & \textbf{5808.7} \\
\bottomrule
\end{tabular}
\begin{tablenotes}[para,flushleft]
\footnotesize \emph{Notes:} RNN/LSTM do not predict $R(t)$; therefore MSE (Restored) is reported as N/A; The test period for the short-period setting corresponds to the last 55 hours and for the long-period setting to the last 96 hours.
\end{tablenotes}
\end{threeparttable}
\end{table}

\begin{table*}[!t]
\centering
\caption{Performance of the generator deployment (synthetic data) with varying travel time ($\delta_t$)
}
\resizebox{\linewidth}{!}{%
\begin{tabular}{llllllllll}
\toprule
\multirow{2}{*}{Model} &
\multicolumn{3}{c}{$\delta_t$ = 1} &
\multicolumn{3}{c}{$\delta_t$ = 5} &
\multicolumn{3}{c}{$\delta_t$ = 10} \\
\cmidrule(lr){2-4} \cmidrule(lr){5-7} \cmidrule(lr){8-10}
& MSE & Cost & Regret & MSE & Cost & Regret & MSE & Cost & Regret \\
\midrule
True Optimal  &  / & 10316.9 (73.9) & 0  & / & 11031.7 (69.0)  & 0  & / & 12473.7 (63.8) & 0 \\
Online (lag = 1) & /      & 63663.2 (3762.5) & 53346.3 (3769.9)  & /      & 18862.8 (965.7) & 7831.1 (916.1)  & /      & 19681.8 (2069.3) & 7208.1 (2121.0) \\
Online (lag = 3) & /      & 84769.4 (122.8) & 74452.6 (7246.8)  & /      & 19816.9 (73.6) & 8785.2 (1216.9) & /      & 38378.6 (3020.2) & 4962.5 (792.7) \\
Neural ODE (Two-stage)  & 9336.0 (975.6)  &  10459.8 (90.6) & 142.9 (90.3)  & 9336.0 (975.6)    &  11506.6 (98.1)  & 474.9 (95.7)    & 9336.0 (975.6)     & 12746.2 (196.7) & 272.4 (207.2) \\ 
Neural ODE (\texttt{GDF})       & \textbf{7204.1 (1804.8)} & 10409.9 (79.7) & \textbf{93.0 (11.8)} & \textbf{7531.5 (1805.8)} & 11452.1 (80.3)  & \textbf{420.3 (24.3)} & \textbf{8907.7 (2645.2)} & 12601.9 (58.1) & \textbf{128.2 (49.3)} \\
\bottomrule
\end{tabular}
}
\label{table:Generator_Distribution_DeltaT}
\end{table*}

\begin{table*}[!t]
\centering
\caption{Performance for the generator deployment (synthetic data) with varying number of warehouses ($W$)
}
\resizebox{\linewidth}{!}{%
\begin{tabular}{llllllllll}
\toprule
\multirow{2}{*}{Model} &
\multicolumn{3}{c}{$W$ = 2} &
\multicolumn{3}{c}{$W$ = 3} &
\multicolumn{3}{c}{$W$ = 4} \\
\cmidrule(lr){2-4} \cmidrule(lr){5-7} \cmidrule(lr){8-10}
& MSE & Cost & Regret & MSE & Cost & Regret & MSE & Cost & Regret \\
\midrule
True Optimal  &  / & 16173.10 & 0  & / & 15639.77 (923.76)  & 0  & / & 12473.7 (63.8) & 0 \\
Online (lag = 1) & /      & 18032.49 (1154.70) & 1859.39 (1154.70)  & /      & 18032.49 (1154.70) & 2392.72 (916.1)  & /      & 18032.49 (1154.70) & 9026.52 (1154.70) \\
Online (lag = 3) & /      & 21500.22 (148.37) & 5327.12 (148.37)  & /      & 21500.22 (148.37) & 5860.45 (135.45) & /      & 21500.22 (148.37) &  5558.79 (148.37) \\
RNN (Two-stage)       & 6356.68 (236.60) & 17904.32 (0.58) & 1731.32 (0.58) & 29537.25 (1304.80) & 17637.40 (462.90) & 1997.63 (460.86) & 29537.25 (1304.80) & 18170.73 (460.86) & 1997.63 (460.86) \\
RNN (\texttt{GDF})       & 30448.27 (1470.67) & 17903.66 (0.00) & 1730.55 (0.00) & 29840.90 (1279.37) &  17637.40 (462.90) & 1997.63 (460.86) & 29839.47 (1279.97) & 18170.73 (460.86) & 1997.63 (460.86) \\
LSTM (Two-stage)       & 10906.76 (7882.75) & 17659.73 (905.46) & 1486.63 (905.46) & 10906.76 (7882.75)  & 17719.59 (983.21)  &  2079.82 (1795.57) & 13277.65 (7321.11)  &18252.93 (1795.57) & 2079.82 (1795.57) \\
LSTM (\texttt{GDF})       & 9777.48 (6663.49) & 18139.86 (1821.40) & 872.22 (755.34) & 13004.73 (8930.83) &  18139.86 (1821.40)  &  2500.09 (1767.27) &  9777.48 (6663.49) & 18275.56 (1762.44) & 2102.46 (1762.44) \\
Neural ODE (Two-stage)  &  1278.43 (27.42)  &  16214.97 (3.70) & 40.51 (1.41)  &  1278.43 (27.42)      &  15680.28 (922.45)  & 39.92 (0.59)    &   1278.43 (27.42)   & 16213.61 (1.41) & 40.51 (1.41) \\ 
Neural ODE (\texttt{GDF})       & \textbf{1047.47 (89.20)} & 16184.52 (0.64) &  \textbf{11.42 (0.64)}  & \textbf{970.80 (32.34)} &  15648.84 (924.81) & \textbf{9.07 (1.06)} & \textbf{976.68 (36.06)} & 16182.01 (1.34) & \textbf{8.91 (1.34)} \\
\bottomrule
\end{tabular}
}
\label{table:Generator_Distribution_multiple_warehouse}
\end{table*}

\subsection{Ablation Study on Neural ODE}
To evaluate the predictive accuracy and physical interpretability of the proposed Neural ODE in \eqref{eq:deterministic sys}, we conduct an ablation study using real-world failure and restoration data from Indianapolis, IN. 
This dataset contains $169,123$ outage records spanning 2004--2024, where each record provides pole-level failure and restoration timestamps.
By aggregating these records to the city level, we construct ground truth trajectories for the failure process $U(t)$ and restoration process $R(t)$, which serve to assess the model’s compartmental interpretation.
We benchmark our approach against standard RNN \cite{rumelhart1986learning} and LSTM \cite{hochreiter1997long} models under identical experimental settings, including enhanced variants trained with all available data preceding the standard training start date in that year (denoted as “extended”). 
Detailed model architectures for the RNN, LSTM, and Neural ODEs are provided in Section~D of the Supplementary Material.

The predictive results in Fig.~\ref{fig:indy-neuralode} indicate that our model captures a distinct ``surge–plateau–restoration'' curve that closely mirrors the observed failure--restoration patterns, while also achieving consistently lower MSE than all baselines across both test periods as shown in Table~\ref{tab:added-experiments}.
Note that the Neural ODE used for short-term forecasting in Table~\ref{tab:added-experiments} contains only two trainable parameters---one governing the failure transmission rate and the other the restoration rate. Despite this highly compact parameterization, our model still delivers superior predictive accuracy compared to all baselines, demonstrating that the underlying physical process is effectively captured by our modeling assumptions. This property is especially advantageous when training data are scarce during hazards.



\subsection{Two Case Studies: Results and Implications}
This section presents the results of \texttt{GDF} on mobile generator deployment and power line undergrounding, evaluated in terms of predictive accuracy and regret. 
All results are evaluated on out-of-sample test sets for both synthetic and real data. 
For synthetic settings, we report both mean and standard deviation in brackets over three runs. 
Results on both synthetic and real datasets demonstrate that \texttt{GDF} improves decision-making compared to conventional Two-stage methods, enabling a more effective response to natural hazards.

\begin{figure}[!t]
\centering
\includegraphics[width=\columnwidth]{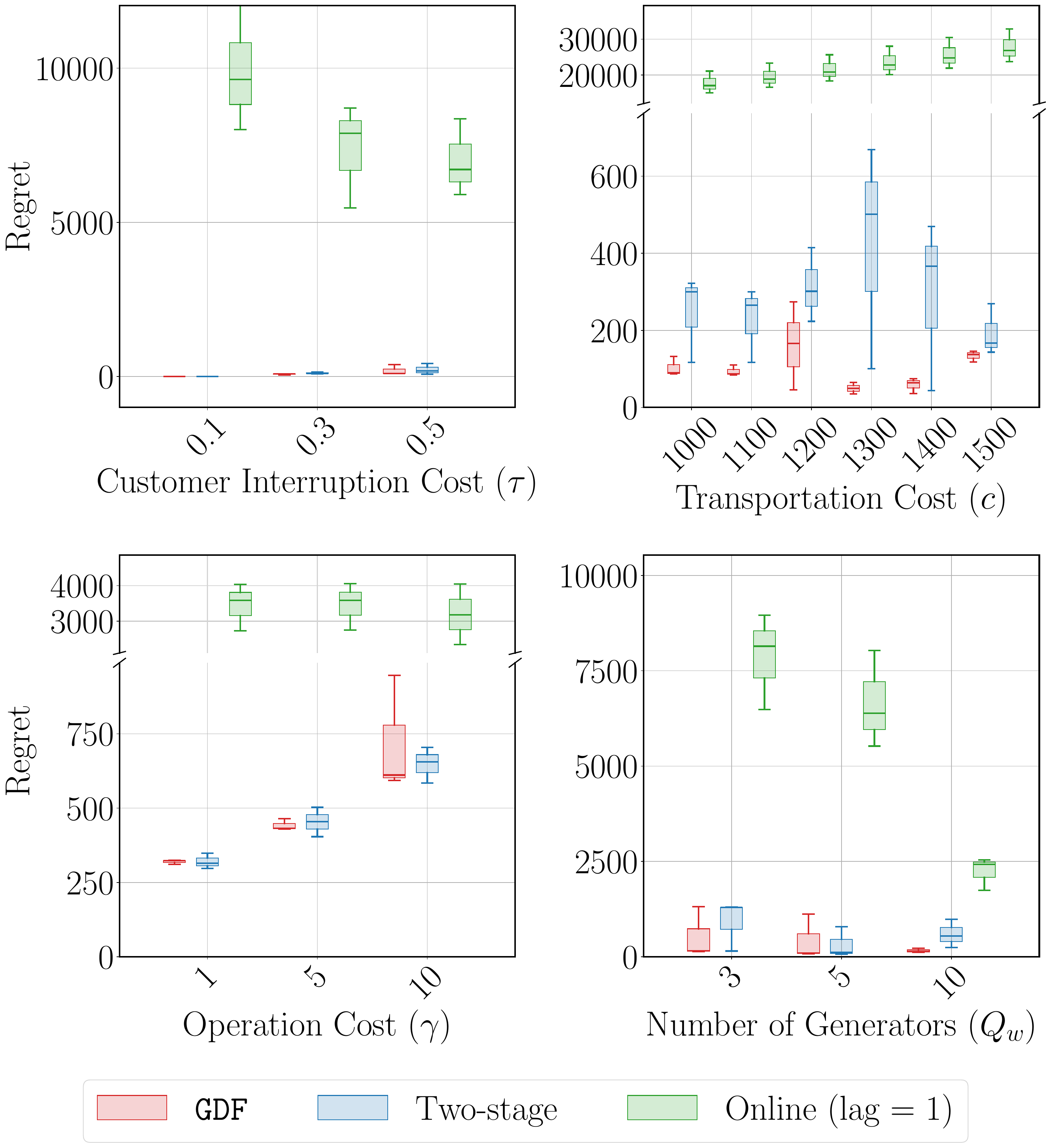}
\caption{Performance Comparison for Synthetic Mobile Generator Deployment: A detailed comparison of regret outcomes for \texttt{GDF}, Two-Stage, and Online methods under varying customer interruption costs ($\tau$), transportation cost factors ($c$), operational costs ($\gamma$), and numbers of generators ($Q_{w}$).}
\label{fig:extensive}
\end{figure}

\subsubsection{Mobile Generator Deployment Problem}
\label{sec:Generator Distribution Problem}
In this problem, we consider $W$ warehouses that store generators. For simplicity, we require each generator to return to a warehouse after completing a task, permitting only transportation between a city and a warehouse, while disallowing direct travel between cities or between warehouses. Furthermore, we impose a stock-balance constraint ensuring that the final number of generators at each warehouse equals to its initial inventory.
Because this is a short-term operational problem, we also benchmark against an online method, a widely used greedy strategy in practice, where decisions are made only after damages are observed, reflecting delays that arise from the absence of predictive insights.

Table~\ref{table:Generator_Distribution_DeltaT} reports the out-of-sample performance on synthetic data with a single warehouse $W=1$ under varying travel times, $\delta_t = 1,5,10$. The results show that our method consistently achieves the lowest regret, and the performance gap between \texttt{GDF} and other baselines widens as $\delta_t$ increases.
Table~\ref{table:Generator_Distribution_multiple_warehouse} presents the out-of-sample performance on the same dataset while varying the number of warehouses, $W = 2,3,4$. As $W$ increases, the problem complexity grows substantially due to the larger set of decision variables. Even under this added complexity, the proposed Neural ODE consistently outperforms all baselines, achieving both stronger predictive accuracy and markedly lower regret.

These results underscore the value of proactive scheduling and early intervention, particularly under longer deployment delays or larger-scale problems. The success of the model can be explained by its behavior shown in Fig.~\ref{fig:diesel}, where the \texttt{GDF} model tends to slightly overestimate outages in the most critical regions initially affected, which enables the reallocation of extra resources to prevent severe disruptions. Although this adjustment in the forecast is subtle, it translates into a substantial reduction in regret. Moreover, the \texttt{GDF} models attain MSE levels comparable to those of the MSE-trained baseline (Table~\ref{table:Generator_Distribution_DeltaT}; Table~\ref{table:Generator_Distribution_multiple_warehouse}), albeit with higher variance. This demonstrates that the gains in decision quality arise not from sacrificing predictive accuracy, but from purposeful and targeted refinements in the forecasts.

Fig.~\ref{fig:extensive} presents additional experiments varying other four key hyperparameters: customer interruption cost ($\tau$), transportation cost ($c$), operation cost ($\gamma$), and the number of generators ($Q_w$). 
We find that the performance gap between \texttt{GDF} and the baselines narrows as all parameters increase, except for transportation cost.
This behavior is expected, as higher system resources or reduced flexibility diminish the need for globally optimized interventions, resulting in less improvement with \texttt{GDF}.


\subsubsection{Power Line Undergrounding}
\label{sec:power-line-undergrounding}


We also investigate power line undergrounding \cite{8278121, 7922545, AbiSamra2013, Shea2018} as a long-term planning application of our framework. The goal is to identify an optimal subset of locations for undergrounding to mitigate risks from future hazards. Unlike the mobile generator deployment problem, this task focuses exclusively on deriving undergrounding decisions by prioritizing the most vulnerable areas based on predicted total damage. Further details of the experimental setup are provided in Section~B of the Supplementary Material.

Table~\ref{table:real} presents the out-of-sample performance on both the Nor’easter, MA, 2018 event and the synthetic dataset. For the real dataset, the proposed \texttt{GDF} model—despite a slightly higher MSE—delivers improved decision quality, achieving the lowest SAIDI and MSE compared to all baselines. On the synthetic dataset, \texttt{GDF} similarly outperforms the Two-stage method in decision quality while maintaining a low MSE. 
Both results demonstrate that \texttt{GDF} offers substantial gains in decision quality compared with the traditional two-stage approach while predictive accuracy remains comparable across models. This underscores the value of proactive scheduling and global decision-focused training for long-term planning tasks.


\begin{table}[!t]
\centering
\caption{Performance of the Power Line Undergrounding}
\resizebox{\linewidth}{!}{%
\begin{tabular}{lllllll}
\toprule
\multirow{2}{*}{Models} & \multicolumn{3}{c}{Nor'easter, MA, 2018} & \multicolumn{3}{c}{Synthetic} \\
\cmidrule(lr){2-4} \cmidrule(lr){5-7}
& {MSE ($\times 10^9)$} & {SAIDI} & {Regret} & {MSE ($\times 10^2)$} & {SAIDI} & {Regret} \\
\midrule
True Optimal    & /      & 0.90 & /   & / & 15.6 & / \\
{RNN (Two-stage)} &  11.3  & 2.20  & 1.3 & 29.5 & 17.2 & 1.6 \\
{RNN ($\texttt{GDF}$)} & 11.3 & 2.20 & 1.3 & 29.8 & 17.2 & 1.6 \\
{LSTM (Two-stage)} & 11.2  & 2.20  & 1.3 & 109.1 & 17.2 & 1.6 \\
{LSTM ($\texttt{GDF}$)} &  11.3 & 2.20 & 1.3 & 130.0 & 15.6 &  \textbf{0} \\
Neural ODE (Two-stage)         & \textbf{5.4} & 1.73 & 0.8   & 13.5 & 16.1 & 0.5 \\
Neural ODE ($\texttt{GDF}$) & 7.1  & 1.56 & \textbf{0.7} & \textbf{13.1} & 15.6 & \textbf{0} \\
\bottomrule
\end{tabular}
}
\label{table:real}
\end{table}

\section{Conclusion}
\label{sec:conclusion}

Extreme hazard events increasingly disrupt power systems, yet conventional predict-then-optimize (PTO) approaches often fall short in high-stakes resilience planning, as small forecasting errors can compound across service units and yield globally suboptimal outcomes. To address this gap, we proposed the predict-all-then-optimize-globally (PATOG) paradigm and developed a global-decision-focused (\texttt{GDF}) Neural ODE model that jointly captures spatio-temporal outage dynamics and system-wide resilience decisions. Methodologically, our approach advances decision-focused learning to multi-unit planning problems by modeling outage progression across regions with Neural ODEs inspired by epidemiological processes. Experiments on both real-world and synthetic datasets demonstrate improved forecast consistency, interpretability, and reduced decision regret, leading to stronger grid resilience. 

We emphasize that PATOG and \texttt{GDF} are broadly applicable to weather-driven challenges such as emergency resource prepositioning, infrastructure hardening, and repair crew staging. 
In particular, the proposed Neural ODE is well suited for continuously evolving, weather-driven hazards, such as storms, hurricanes, wildfires, and heatwaves. 
This flexibility across hazard contexts underscores its value for utilities and policymakers seeking proactive, region-specific resilience planning.
Future work may extend this framework by incorporating transportation and travel-time uncertainties into the decision layer to better capture logistical delays and further enhance its operational realism.

\section*{Acknowledgments}
We sincerely appreciate Dr. James Kotary for his valuable insights on implementing quadratic regularization in decision-focused learning and for enhancing the literature review.

\bibliographystyle{IEEEtran}
\bibliography{ref}

@misc{gurobi,
  author = {{Gurobi Optimization, LLC}},
  title = {{Gurobi Optimizer Reference Manual}},
  year = 2024,
  url = "https://www.gurobi.com"
}

@article{rumelhart1986learning,
  title={Learning representations by back-propagating errors},
  author={Rumelhart, David E and Hinton, Geoffrey E and Williams, Ronald J},
  journal={Nature},
  volume={323},
  number={6088},
  pages={533--536},
  year={1986},
  publisher={Nature Publishing Group},
  doi={10.1038/323533a0}
}

@article{hochreiter1997long,
  title={Long short-term memory},
  author={Hochreiter, Sepp and Schmidhuber, J{\"u}rgen},
  journal={Neural Computation},
  volume={9},
  number={8},
  pages={1735--1780},
  year={1997},
  publisher={MIT Press},
  doi={10.1162/neco.1997.9.8.1735}
}

@article{gould2021deep,
  title={Deep declarative networks},
  author={Gould, Stephen and Hartley, Richard and Campbell, Dylan},
  journal={{IEEE} Transactions on Pattern Analysis and Machine Intelligence},
  volume={44},
  number={8},
  pages={3988--4004},
  year={2021},
  publisher={IEEE}
}

@inproceedings{kotary2021end,
  title     = {End-to-End Constrained Optimization Learning: A Survey},
  author    = {Kotary, James and Fioretto, Ferdinando and Van Hentenryck, Pascal and Wilder, Bryan},
  booktitle = {Proceedings of the Thirtieth International Joint Conference on
               Artificial Intelligence, {IJCAI-21}},
  pages     = {4475--4482},
  year      = {2021},
  doi       = {10.24963/ijcai.2021/610},
  url       = {https://doi.org/10.24963/ijcai.2021/610},
}

@standard{ieee1366_2022,
  title        = {{IEEE Guide for Electric Power Distribution Reliability Indices}},
  organization = {IEEE},
  institution  = {Institute of Electrical and Electronics Engineers},
  number       = {IEEE Std 1366-2022 (Revision of IEEE Std 1366-2012)},
  pages        = {1--44},
  year         = {2022},
  doi          = {10.1109/IEEESTD.2022.9747211}
}

@misc{USCensusBureau2017ACS,
  author       = {{U.S. Census Bureau}},
  title        = {American Community Survey 1-Year Estimates, Data Profiles: Massachusetts},
  year         = 2017,
  url          = {https://www.census.gov/acs/www/data/data-tables-and-tools/data-profiles/2017/},
  note         = {Accessed: 2025-02-23}
}

@techreport{maema2020power,
  title        = {Massachusetts Power Outages},
  year         = {2020},
  institution  = {Massachusetts Emergency Management Agency},
  address      = {Massachusetts, USA}
}

@misc{torchdiffeq,
	author={Chen, Ricky T. Q.},
	title={torchdiffeq},
	year={2018},
	url={https://github.com/rtqichen/torchdiffeq},
}

@article{zhang2024recurrent,
  title={Recurrent Neural Goodness-of-Fit Test for Time Series},
  author={Zhang, Aoran and Zhou, Wenbin and Xie, Liyan and Zhu, Shixiang},
  journal={arXiv preprint arXiv:2410.13986},
  year={2024}
}

@INPROCEEDINGS{LSTM,
  author={Maniamfu, Pavodi and Kameyama, Keisuke},
  booktitle={2023 19th {IEEE} International Colloquium on Signal Processing and Its Applications (CSPA)}, 
  title={LSTM-based Forecasting using Policy Stringency and Time-varying Parameters of the SIR Model for COVID-19}, 
  year={2023},
  volume={},
  number={},
  pages={111-116},
  doi={10.1109/CSPA57446.2023.10087773}
}

@article{zhu2021quantifying,
author = {Zhu, Shixiang and Yao, Rui and Xie, Yao and Qiu, Feng and Qiu, Yueming (Lucy) and Wu, Xuan},
title = {Quantifying Grid Resilience Against Extreme Weather Using Large-Scale Customer Power Outage Data},
journal = {INFORMS Journal on Data Science},
volume = {0},
number = {0},
pages = {null},
year = {0},
doi = {10.1287/ijds.2023.0017},

URL = { 
    
        https://doi.org/10.1287/ijds.2023.0017
    
    

},
eprint = { 
    
        https://doi.org/10.1287/ijds.2023.0017
    
    

}
}

@incollection{Handmer2012,
    title={Changes in impacts of climate extremes: human systems and ecosystems},
    author={Handmer, John and Honda, Yasushi and Kundzewicz, Zbigniew W and Arnell, Nigel and Benito, Gerardo and Hatfield, Jerry and Mohamed, Ismail Fadl and Peduzzi, Pascal and Wu, Shaohong and Sherstyukov, Boris and others},
    booktitle={Managing the risks of extreme events and disasters to advance climate change adaptation special report of the intergovernmental panel on climate change},
    pages={231--290},
    year={2012},
    publisher={Intergovernmental Panel on Climate Change}
}

@article{Kenward2014,
    title={Blackout: Extreme weather climate change and power outages},
    author={Kenward, Alyson and Raja, Urooj},
    journal={Climate central},
    volume={10},
    pages={1--23},
    year={2014}
}

@inproceedings{mandi2020smart,
  title={Smart predict-and-optimize for hard combinatorial optimization problems},
  author={Mandi, Jayanta and Stuckey, Peter J and Guns, Tias and others},
  booktitle={Proceedings of the AAAI Conference on Artificial Intelligence},
  volume={34},
  number={02},
  pages={1603--1610},
  year={2020}
}

@article{mandi_decision-focused_2024,
	title = {Decision-{Focused} {Learning}: {Foundations}, {State} of the {Art}, {Benchmark} and {Future} {Opportunities}},
	volume = {80},
	copyright = {Copyright (c) 2024 Journal of Artificial Intelligence Research},
	issn = {1076-9757},
	shorttitle = {Decision-{Focused} {Learning}},
	url = {https://www.jair.org/index.php/jair/article/view/15320},
	doi = {10.1613/jair.1.15320},
	language = {en},
	urldate = {2024-09-25},
	journal = {Journal of Artificial Intelligence Research},
	author = {Mandi, Jayanta and Kotary, James and Berden, Senne and Mulamba, Maxime and Bucarey, Victor and Guns, Tias and Fioretto, Ferdinando},
	month = aug,
	year = {2024},
	pages = {1623--1701},
}

@article{vlastelica2019differentiation,
  title={Differentiation of blackbox combinatorial solvers},
  author={Vlastelica, Marin and Paulus, Anselm and Musil, V{\'\i}t and Martius, Georg and Rol{\'\i}nek, Michal},
  journal={arXiv preprint arXiv:1912.02175},
  year={2019}
}

@article{HoustonChronicle2025Grid,
  author       = "{Houston Chronicle Staff}",
  title        = "{Entergy Texas invests \$137 million in grid improvements to withstand extreme weather}",
  journal      = "Houston Chronicle",
  year         = 2025,
  url          = "https://www.houstonchronicle.com/business/energy/article/entergy-grid-improvements-extreme-weather-20034162.php",
  note         = "[Online; accessed 14-February-2025]"
}

@article{CNN2025Wildfires,
  author       = "{Majlie de Puy Kamp, Curt Devine, Casey Tolan, Blake Ellis, Melanie Hicken, Rob Kuznia, Scott Glover, Yahya Abou-Ghazala, Audrey Ash, and Nelli Black}",
  title        = "{No ‘water system in the world’ could have handled the LA fires. How the region could have minimized the damage}",
  journal      = "CNN",
  year         = 2025,
  url          = "https://www.cnn.com/2025/01/10/us/los-angeles-wildfires-water-system-analysis/index.html",
  note         = "[Online; accessed 14-February-2025]"
}

@misc{wiki:2019_CA_power_shutoffs,
  author       = "{Wikipedia contributors}",
  title        = "{2019 California power shutoffs --- Wikipedia{,} The Free Encyclopedia}",
  year         = "2019",
  url          = "https://en.wikipedia.org/wiki/2019_California_power_shutoffs",
  note         = "[Online; accessed 14-February-2025]"
}

@inproceedings{wilder2019melding,
  title={Melding the data-decisions pipeline: Decision-focused learning for combinatorial optimization},
  author={Wilder, Bryan and Dilkina, Bistra and Tambe, Milind},
  booktitle={Proceedings of the AAAI Conference on Artificial Intelligence},
  volume={33},
  number={01},
  pages={1658--1665},
  year={2019}
}

@INPROCEEDINGS{8668633,
  author={Deshmukh, Rashmi and Kalage, Amol},
  booktitle={2018 {IEEE} Global Conference on Wireless Computing and Networking (GCWCN)}, 
  title={Optimal Placement and Sizing of Distributed Generator in Distribution System Using Artificial Bee Colony Algorithm}, 
  year={2018},
  volume={},
  number={},
  pages={178-181},
  doi={10.1109/GCWCN.2018.8668633}}

@ARTICLE{7922545,
  author={Panteli, Mathaios and Trakas, Dimitris N. and Mancarella, Pierluigi and Hatziargyriou, Nikos D.},
  journal={Proceedings of the {IEEE}}, 
  title={Power Systems Resilience Assessment: Hardening and Smart Operational Enhancement Strategies}, 
  year={2017},
  volume={105},
  number={7},
  pages={1202-1213},
  doi={10.1109/JPROC.2017.2691357}}

@INPROCEEDINGS{8273985,
  author={Qiu, Jing and Reedman, Luke J. and Dong, Zhao Yang and Meng, Ke and Tian, Huiqiao and Zhao, Junhua},
  booktitle={2017 {IEEE} Power \& Energy Society General Meeting}, 
  title={Network reinforcement for grid resiliency under extreme events}, 
  year={2017},
  volume={},
  number={},
  pages={1-5},
  doi={10.1109/PESGM.2017.8273985}}

@INPROCEEDINGS{990600,
  author={Aydin, H. and Melhem, R. and Mosse, D. and Mejia-Alvarez, P.},
  booktitle={Proceedings 22nd {IEEE} Real-Time Systems Symposium (RTSS 2001) (Cat. No.01PR1420)}, 
  title={Dynamic and aggressive scheduling techniques for power-aware real-time systems}, 
  year={2001},
  volume={},
  number={},
  pages={95-105},
  doi={10.1109/REAL.2001.990600}}

@INPROCEEDINGS{5963580,
  author={Bu, Shengrong and Yu, F. Richard and Liu, Peter X. and Zhang, Peng},
  booktitle={2011 {IEEE} International Conference on Communications Workshops (ICC)}, 
  title={Distributed Scheduling in Smart Grid Communications with Dynamic Power Demands and Intermittent Renewable Energy Resources}, 
  year={2011},
  volume={},
  number={},
  pages={1-5},
  doi={10.1109/iccw.2011.5963580}}

@INPROCEEDINGS{9117454,
  author={Ahmed, Nabil A. and AlHajri, Mohamad F.},
  booktitle={2019 {IEEE} 6th International Conference on Engineering Technologies and Applied Sciences (ICETAS)}, 
  title={Distributed Generators Optimal Placement and Sizing in Power Systems}, 
  year={2019},
  volume={},
  number={},
  pages={1-5},
  doi={10.1109/ICETAS48360.2019.9117454}}

@manual{AbiSamra2013,
    title  = "Hardening the System",
    author = "Abi-Samra, Nicholas and Willis, Lee and Moon, Marvin",
    url    = "https://www.tdworld.com/vegetation-management/article/20962556/hardening-the-system",
    year   = "2013"
}

@ARTICLE{10855832,
  author={Massaoudi, Mohamed and Ez Eddin, Maymouna and Ghrayeb, Ali and Abu-Rub, Haitham and Refaat, Shady S.},
  journal={{IEEE} Open Access Journal of Power and Energy}, 
  title={Advancing Coherent Power Grid Partitioning: A Review Embracing Machine and Deep Learning}, 
  year={2025},
  volume={12},
  number={},
  pages={59-75},
  doi={10.1109/OAJPE.2025.3535709}}

@manual{Shea2018,
    title  = "Hardening the Grid: How States Are Working to Establish a Resilient and Reliable Electric System",
    author = "Daniel Shea",
    year   = "2018",
    note   = "Available online: \\url{https://www.ncsl.org/research/energy/hardening-the-grid-how-states-are-working-to-establish-a-resilient-and-reliable-electric-system.aspx} [Accessed: February 14, 2025]"
}

@inproceedings{Fioretto:ICLR25,
  title = {Learning To Solve Differential Equation Constrained Optimization Problems},
  author = {Vito, Vincenzo Di and Mohammadian, Mostafa and Baker, Kyri and Fioretto, Ferdinando},
  year = {2025},
  booktitle = {International Conference on Learning Representations},
  volume = {13},
  doi = {10.48550/arXiv.2410.01786},
}

@article{fernandez2022causal,
  title={Causal decision making and causal effect estimation are not the same… and why it matters},
  author={Fern{\'a}ndez-Lor{\'\i}a, Carlos and Provost, Foster},
  journal={INFORMS Journal on Data Science},
  volume={1},
  number={1},
  pages={4--16},
  year={2022},
  publisher={INFORMS}
}

@article{elmachtoub_smart_2022,
	title = {Smart “{Predict}, then {Optimize}”},
	volume = {68},
	issn = {0025-1909},
	url = {https://pubsonline.informs.org/doi/10.1287/mnsc.2020.3922},
	doi = {10.1287/mnsc.2020.3922},
	number = {1},
	urldate = {2024-09-25},
	journal = {Management Science},
	author = {Elmachtoub, Adam N. and Grigas, Paul},
	month = jan,
	year = {2022},
	note = {Publisher: INFORMS},
	pages = {9--26},
}

@INPROCEEDINGS{9646114,
  author={Qian, Cheng and Wang, Aiyuan},
  booktitle={2021 {IEEE} 4th Student Conference on Electric Machines and Systems (SCEMS)}, 
  title={Power Grid Disturbance Prediction and Analysis Method Based on {SIR} Model}, 
  year={2021},
  volume={},
  number={},
  pages={1-5},
  doi={10.1109/SCEMS52239.2021.9646114}}

@article{berthet2020learning,
  title={Learning with differentiable pertubed optimizers},
  author={Berthet, Quentin and Blondel, Mathieu and Teboul, Olivier and Cuturi, Marco and Vert, Jean-Philippe and Bach, Francis},
  journal={Advances in neural information processing systems},
  volume={33},
  pages={9508--9519},
  year={2020}
}

@misc{noaa_hrrr,
  author       = {{National Oceanic and Atmospheric Administration}},
  title        = {High-Resolution Rapid Refresh ({HRRR}) Model},
  year         = {2024},
  howpublished = {\url{https://rapidrefresh.noaa.gov/hrrr/}},
  note         = {Accessed: 2024-11-11}
}

@ARTICLE{9955492,
  author={},
  journal={{IEEE} Std 1366-2022 (Revision of {IEEE} Std 1366-2012)}, 
  title={{IEEE} Guide for Electric Power Distribution Reliability Indices}, 
  year={2022},
  volume={},
  number={},
  pages={1-44},
  doi={10.1109/IEEESTD.2022.9955492}}

@article{QIN2024752,
title = {The generator distribution problem for base stations during emergency power outage: A branch-and-price-and-cut approach},
journal = {European Journal of Operational Research},
volume = {318},
number = {3},
pages = {752-767},
year = {2024},
issn = {0377-2217},
doi = {https://doi.org/10.1016/j.ejor.2024.06.007},
url = {https://www.sciencedirect.com/science/article/pii/S0377221724004351},
author = {Hu Qin and Anton Moriakin and Gangyan Xu and Jiliu Li}
}

@article{kotary2023folded,
      title={Backpropagation of Unrolled Solvers with Folded Optimization}, 
      author={James Kotary and My H. Dinh and Ferdinando Fioretto},
      year={2023},
      journal={arXiv preprint arXiv:2301.12047}
}

@article{wilder2019end,
  title={End to end learning and optimization on graphs},
  author={Wilder, Bryan and Ewing, Eric and Dilkina, Bistra and Tambe, Milind},
  journal={Advances in Neural Information Processing Systems},
  volume={32},
  year={2019}
}

@inproceedings{cvx,
  author={Agrawal, A. and Amos, B. and Barratt, S. and Boyd, S. and Diamond, S. and Kolter, Z.},
  title={Differentiable Convex Optimization Layers},
  booktitle={Advances in Neural Information Processing Systems},
  year={2019},
}

@InProceedings{amos2017optnet,
  title = {{O}pt{N}et: Differentiable Optimization as a Layer in Neural Networks},
  author = {Brandon Amos and J. Zico Kolter},
  booktitle = {Proceedings of the 34th International Conference on Machine Learning},
  pages = {136--145},
  year = {2017},
  volume = {70},
  series = {Proceedings of Machine Learning Research},
  publisher ={PMLR},
}

@article{kosma2023neural,
  title={Neural Ordinary Differential Equations for Modeling Epidemic Spreading},
  author={Kosma, Chrysoula and Nikolentzos, Giannis and Panagopoulos, George and Steyaert, Jean-Marc and Vazirgiannis, Michalis},
  journal={Transactions on Machine Learning Research},
  year={2023}
}

@INPROCEEDINGS{10711393,
  author={Guo, Wangyi and Xu, Zhanbo and Zhou, Zhequn and Liu, Jinhui and Wu, Jiang and Zhao, Haoming and Guan, Xiaohong},
  booktitle={2024 {IEEE} 20th International Conference on Automation Science and Engineering (CASE)}, 
  title={Integrating end-to-end prediction-with-optimization for distributed hydrogen energy system scheduling*}, 
  year={2024},
  volume={},
  number={},
  pages={2774-2779},
  doi={10.1109/CASE59546.2024.10711393}}

@ARTICLE{7080837,
  author={Giri, Jay},
  journal={{IEEE} Power and Energy Technology Systems Journal}, 
  title={Proactive Management of the Future Grid}, 
  year={2015},
  volume={2},
  number={2},
  pages={43-52},
  doi={10.1109/JPETS.2015.2408212}}

@INPROCEEDINGS{328395,
  author={Kramer, S.R. and Rodenbaugh, T.J. and Conroy, M.W.},
  booktitle={Proceedings of {IEEE}/PES Transmission and Distribution Conference}, 
  title={The use of trenchless technologies for transmission and distribution projects}, 
  year={1994},
  volume={},
  number={},
  pages={302-308},
  doi={10.1109/TDC.1994.328395}}

@ARTICLE{8278121,
  author={Fairley, Peter},
  journal={{IEEE} Spectrum}, 
  title={Utilities bury transmission lines}, 
  year={2018},
  volume={55},
  number={2},
  pages={9-10},
  doi={10.1109/MSPEC.2018.8278121}}

@ARTICLE{9220164,
  author={Muhs, John W. and Parvania, Masood and Shahidehpour, Mohammad},
  journal={{IEEE} Open Access Journal of Power and Energy}, 
  title={Wildfire Risk Mitigation: A Paradigm Shift in Power Systems Planning and Operation}, 
  year={2020},
  volume={7},
  number={},
  pages={366-375},
  doi={10.1109/OAJPE.2020.3030023}}

@inproceedings{chen2018neural,
author = {Chen, Ricky T. Q. and Rubanova, Yulia and Bettencourt, Jesse and Duvenaud, David},
title = {Neural ordinary differential equations},
year = {2018},
publisher = {Curran Associates Inc.},
address = {Red Hook, NY, USA},
booktitle = {Proceedings of the 32nd International Conference on Neural Information Processing Systems},
pages = {6572–6583},
numpages = {12},
location = {Montr\'{e}al, Canada},
series = {NIPS'18}
}

@ARTICLE{7752978,
  author={Eskandarpour, Rozhin and Khodaei, Amin},
  journal={{IEEE} Transactions on Power Systems}, 
  title={Machine Learning Based Power Grid Outage Prediction in Response to Extreme Events}, 
  year={2017},
  volume={32},
  number={4},
  pages={3315-3316},
  doi={10.1109/TPWRS.2016.2631895}}

@INPROCEEDINGS{9160513,
  author={Macaš, Martin and Orlando, Sergio and Costea, Stefan and Novák, Petr and Chumak, Oleksiy and Kadera, Petr and Kopejtko, Petr},
  booktitle={2020 {IEEE} International Conference on Environment and Electrical Engineering and 2020 {IEEE} Industrial and Commercial Power Systems Europe (EEEIC / I\&CPS Europe)}, 
  title={Impact of forecasting errors on microgrid optimal power management}, 
  year={2020},
  volume={},
  number={},
  pages={1-6},
  doi={10.1109/EEEIC/ICPSEurope49358.2020.9160513}}

@article{reuters2025wildfires,
  author  = {Reuters},
  title   = {Californian utility SoCal Edison shuts power to over 114,000 customers due to wildfire},
  journal = {Reuters},
  year    = {2025},
  month   = {January},
  day     = {8},
  url     = {https://www.reuters.com/business/energy/californian-utility-socal-edison-shuts-power-over-114000-customers-due-wildfire-2025-01-08/},
  note    = {Accessed: 2025-02-15}
}

@article{jacquillat2024branch,
  author = {Alexandre Jacquillat and Michael Lingzhi Li and Martin Ramé and Kai Wang},
  title = {Branch-and-Price for Prescriptive Contagion Analytics},
  journal = {Operations Research},
  year = {2024},
  volume = {Ahead of Print},
  doi = {10.1287/opre.2023.0308},
  url = {https://doi.org/10.1287/opre.2023.0308}
}

@misc{wang2025gendfldecisionfocusedgenerativelearning,
      title={Gen-DFL: Decision-Focused Generative Learning for Robust Decision Making}, 
      author={Prince Zizhuang Wang and Jinhao Liang and Shuyi Chen and Ferdinando Fioretto and Shixiang Zhu},
      year={2025},
      eprint={2502.05468},
      archivePrefix={arXiv},
      primaryClass={cs.LG},
      url={https://arxiv.org/abs/2502.05468}, 
}

@article{reuters2024hurricane,
  author  = {Reuters},
  title   = {Over 1.3 million Florida customers without power due to Hurricane Milton},
  journal = {Reuters},
  year    = {2024},
  month   = {October},
  day     = {10},
  url     = {https://www.reuters.com/business/energy/over-13-million-florida-customers-without-power-due-hurricane-milton-2024-10-10/},
  note    = {Accessed: 2025-02-15}
}

@ARTICLE{10479971,
  author={Rajkumar, Vetrivel S. and Ştefanov, Alexandru and Rueda Torres, José Luis and Palensky, Peter},
  journal={{IEEE} Transactions on Industrial Informatics}, 
  title={Dynamical Analysis of Power System Cascading Failures Caused by Cyber Attacks}, 
  year={2024},
  volume={20},
  number={6},
  pages={8807-8817},
  doi={10.1109/TII.2024.3372024}}

@INPROCEEDINGS{chen2025uncertainty,
  author={Chen, Shuyi and Ding, Kaize and Zhu, Shixiang},
  booktitle={ICASSP 2025 - 2025 IEEE International Conference on Acoustics, Speech and Signal Processing (ICASSP)}, 
  title={Uncertainty-Aware Robust Learning on Noisy Graphs}, 
  year={2025},
  volume={},
  number={},
  pages={1-5},
  doi={10.1109/ICASSP49660.2025.10888672}}

@INPROCEEDINGS{6672826,
  author={Ratha, Anubhav and Iggland, Emil and Andersson, Göran},
  booktitle={2013 {IEEE} Power \& Energy Society General Meeting}, 
  title={Value of Lost Load: How much is supply security worth?}, 
  year={2013},
  volume={},
  number={},
  pages={1-5},
  doi={10.1109/PESMG.2013.6672826}}

@article{zhu2022distributionally,
  title={Distributionally robust weighted k-nearest neighbors},
  author={Zhu, Shixiang and Xie, Liyan and Zhang, Minghe and Gao, Rui and Xie, Yao},
  journal={Advances in Neural Information Processing Systems},
  volume={35},
  pages={29088--29100},
  year={2022}
}

@ARTICLE{9803820,
  author={Xie, Le and Zheng, Xiangtian and Sun, Yannan and Huang, Tong and Bruton, Tony},
  journal={Proceedings of the IEEE}, 
  title={Massively Digitized Power Grid: Opportunities and Challenges of Use-Inspired AI}, 
  year={2023},
  volume={111},
  number={7},
  pages={762-787},
  doi={10.1109/JPROC.2022.3175070}}
\appendix

\subsection{Ablation Study on $\lambda$}
\label{sec:alb}
To evaluate the impact of the prediction error weight $\lambda$ in (6) on balancing prediction accuracy and decision quality, we conducted an ablation study with various $\lambda$ values on synthetic dataset. Results in Table~\ref{table:alb} indicate that a lower $\lambda$ shifts the model’s focus toward decision quality, reducing decision regret by aligning predictions with resilience goals, albeit with a slight trade-off in MSE, compared to Two-stage method. We note that such a design offers better flexibility and interpretability for the \texttt{GDF}-trained decision-making models.

\begin{table}[ht]
\centering
\caption{Effect of $\lambda$ on Mobile Generator Deployment Problem.}
\resizebox{0.9\linewidth}{!}{%
\begin{tabular}{lccccc}
\toprule
\multirow{2}{*}{Metric} & \multicolumn{3}{c}{$\lambda$} & \multicolumn{2}{c}{Baselines} \\
\cmidrule(lr){2-4} \cmidrule(lr){5-6}
                      & 0  & 1   & 10   & Two-stage & Online \\
\midrule
MSE ($\times 10^3$)    & 9.7 & 9.5 &  9.6  &  9.3 & /      \\
Cost         & 6668.7 & 6609.5 & 6611.8  & 6631.7 &  8163.7   \\
Regret         & 153.0 & 135.7 &  138.0  &  155.8 &   1648.1   \\
\bottomrule
\end{tabular}
}
\label{table:alb}
\end{table}

\subsection{Power Line Undergrounding Problem}
\label{exp:power hardening}
Power line undergrounding is a grid-hardening measure against extreme meteorological events (such as hurricanes and heavy snowfalls) \cite{8278121, 7922545,AbiSamra2013,Shea2018}. Although effective, it involves substantial costs for the authority and causes significant disruptions to local communities \cite{9220164,328395}. The objective of the power line undergrounding problem is to select an optimal subset of locations for underground interventions under budget constraints \cite{7922545, 
AbiSamra2013, Shea2018}, in anticipation of an incoming hazard.

To formalize the decision-making problem, let $x_k$ be a binary variable indicating whether city~$k$ is selected for undergrounding. 
With $K$ cities in total, the decision vector is $\boldsymbol{x}=[x_1,\dots,x_K]^\top$, and $\hat{\mathbf{S}}$ denotes the predicted outage states. 
We then define:
\begin{equation}
\label{eq:opt_power}
\begin{aligned}
\min_{\boldsymbol{x}} \quad & g(\boldsymbol{x}, \hat{\mathbf{S}}),\\
\text{s.t.}\quad
  & \sum_{k=1}^{K} x_k \,\leq\, C,\\
  & x_k \,\in\, \{0,1\},\quad k = 1,\dots,K,
\end{aligned}
\end{equation}
where $g(\boldsymbol{x}, \hat{\mathbf{S}})$ is the decision loss that quantifies the impact of outages given the chosen undergrounding plan $\boldsymbol{x}$.


We adopt the System Average Interruption Duration Index (SAIDI)~\cite{9955492} to measure how outages affect the population. 
Let $Y_k(t)$ be the (true) number of outages at city $k$ and time $t$, and $N_k$ be the total number of customers in city~$k$. 
Since undergrounding is assumed fully effective, a city $k$ with $x_k=1$ incurs no further outages from the event. 
Hence, the decision loss is:
\begin{equation}
\label{eq:saidi}
g(\boldsymbol{x}, \mathbf{S}) 
 \;=\; \frac{1}{K}\,\sum_{k=1}^{K} 
  \frac{1}{N_k} \int_{0}^{\infty} 
    \Bigl[\bigl(1 - x_k\bigr)\,Y_k(t)\Bigr]\,
  dt.
\end{equation}
The optimal solution $\boldsymbol{x}^*$ to~\eqref{eq:opt_power} is then the subset of cities to be undergrounded in order to minimize the total outage impact. 
Its performance is evaluated via $g(\boldsymbol{x}^*, \mathbf{S})$ using true outage data $\mathbf{S}$.



\subsection{Implementation Details of \texttt{GDF}}
\label{append:implement}

To get the gradient of (7), let matrix $H$ encode coefficients for all the linear constraints $\quad H\boldsymbol{x} \leq \boldsymbol{a}$, and ${\xi^i}$ represents a reformulation of the ground truth cost factors derived from $\mathbf{S}^i$, such that $g(\boldsymbol{x},\mathbf{S}^i) = {\xi^i}^T \boldsymbol{x} + \boldsymbol{b}$. Using the KKT conditions of the Lagrangian of the problem, the gradient of the QP in (7) is:
\begin{equation}
\nabla_{\theta} L_{\text{QP}} = {\xi^i}^T K^{-1} \nabla_{\theta} \hat{\mathbf{S}}^i, \quad
K =
\begin{bmatrix}
2\rho_x I & H^T \\
H & 0
\end{bmatrix},
\label{eq:event_decision_gradient}
\end{equation}

And $\nabla_{\beta, \gamma} \hat{\mathbf{S}}^i$ can obtained via backpropagation through the neural ODE model parameters $\{\theta_U, \theta_R\}$ \cite{chen2018neural}.
This gradient aims to improve decision quality across all cities and all events $i \in \{1, \dots, I\}$. Therefore, we refer to it as \emph{decision-focused gradient}.

To be more specific about derivation of \eqref{eq:event_decision_gradient}, the optimal solution $\boldsymbol{x}^*$ must satisfy the KKT conditions. We define the Lagrangian:
\begin{equation}
\mathcal{L}(\boldsymbol{x}, \lambda) = g(\boldsymbol{x}, \mathbf{S}) + \rho \|\boldsymbol{x}\|_2^2 + \lambda^T (H \boldsymbol{x} - \boldsymbol{a}).
\end{equation}

The stationarity of $x^*$ for optimality gives:
\begin{equation}
\nabla_x g(\boldsymbol{x}, \mathbf{S}) + 2\rho \boldsymbol{x} + H^T \lambda = 0.
\end{equation}

Since the optimal decision $\boldsymbol{x}^*$ satisfies the above KKT system, we can apply implicit differentiation. Taking the total derivative with respect to the predicted system state $\hat{\mathbf{S}}$:

\begin{equation}
\begin{bmatrix}
\nabla^2_{xx} \mathcal{L} & H^T \\
H & 0
\end{bmatrix}
\begin{bmatrix}
\frac{d\boldsymbol{x}^*}{d\hat{\mathbf{S}}} \\
\frac{d\lambda}{d\hat{\mathbf{S}}}
\end{bmatrix}
=
\begin{bmatrix}
\frac{d (-\nabla_x g(\boldsymbol{x}, \mathbf{S}))}{d\hat{\mathbf{S}}} \\
0
\end{bmatrix}.
\label{eq:kkt_diff}
\end{equation}

where
\begin{equation}
\frac{d\boldsymbol{x}^*}{d\hat{\mathbf{S}}} = - K^{-1} \frac{d \nabla_x g}{d\hat{\mathbf{S}}},
\end{equation}
and the KKT matrix $K$ is defined as:
\begin{equation}
K =
\begin{bmatrix}
2\rho I & H^T \\
H & 0
\end{bmatrix}.
\end{equation}
since $\nabla^2_{xx} \mathcal{L} = 2\rho I$ etc.

This allows us to compute the gradient of the loss function with respect to model parameters:
\begin{equation}
\nabla_{\theta} L_{\text{QP}} = {\xi^i}^T K^{-1} \nabla_{\theta} \hat{\mathbf{S}}^i.
\end{equation}

In practice, we regularize the \texttt{GDF} loss with prediction-focused gradient,
as the prediction error is localized and requires fine-grained information for each individual sample unit. 
Specifically, we construct mini-batches $\mathcal{B} \subset \mathcal{D}$ from the training dataset $\mathcal{D} = \{\boldsymbol{z}_k^i, y_k^i(t)\}$.

The neural ODE model generates forecasts $\hat{\mathbf{S}}_{\mathcal{B}} = f_{\beta, \gamma}(\boldsymbol{z}_{\mathcal{B}})$ for each sample in the batch. From these forecasts, we extract the predicted values $\hat{Y}_{\mathcal{B}}$, which are then used to compute the MSE loss:
\begin{equation}
L_{\text{MSE}} = \sum_{\mathcal{B} \subset \mathcal{D}}\frac{1}{|\mathcal{B}|} \sum_{(i,k,t)\in \mathcal{B}} \bigl(y_k^i(t) - \hat{Y}_k^i(t)\bigr)^2.
\end{equation}

Finally, for each epoch, the model parameters $\theta$ is updated using a combination of both loss gradients, balanced by a hyperparameter $\lambda$:
\begin{equation}
\nabla_{\theta} L_{\text{total}} = \nabla_{\theta} {L_{\text{QP}}} + \lambda \nabla_{\theta} L_{\text{MSE}}.
\end{equation}

\begin{figure}[!t]
\centering
\includegraphics[width=0.9\columnwidth]{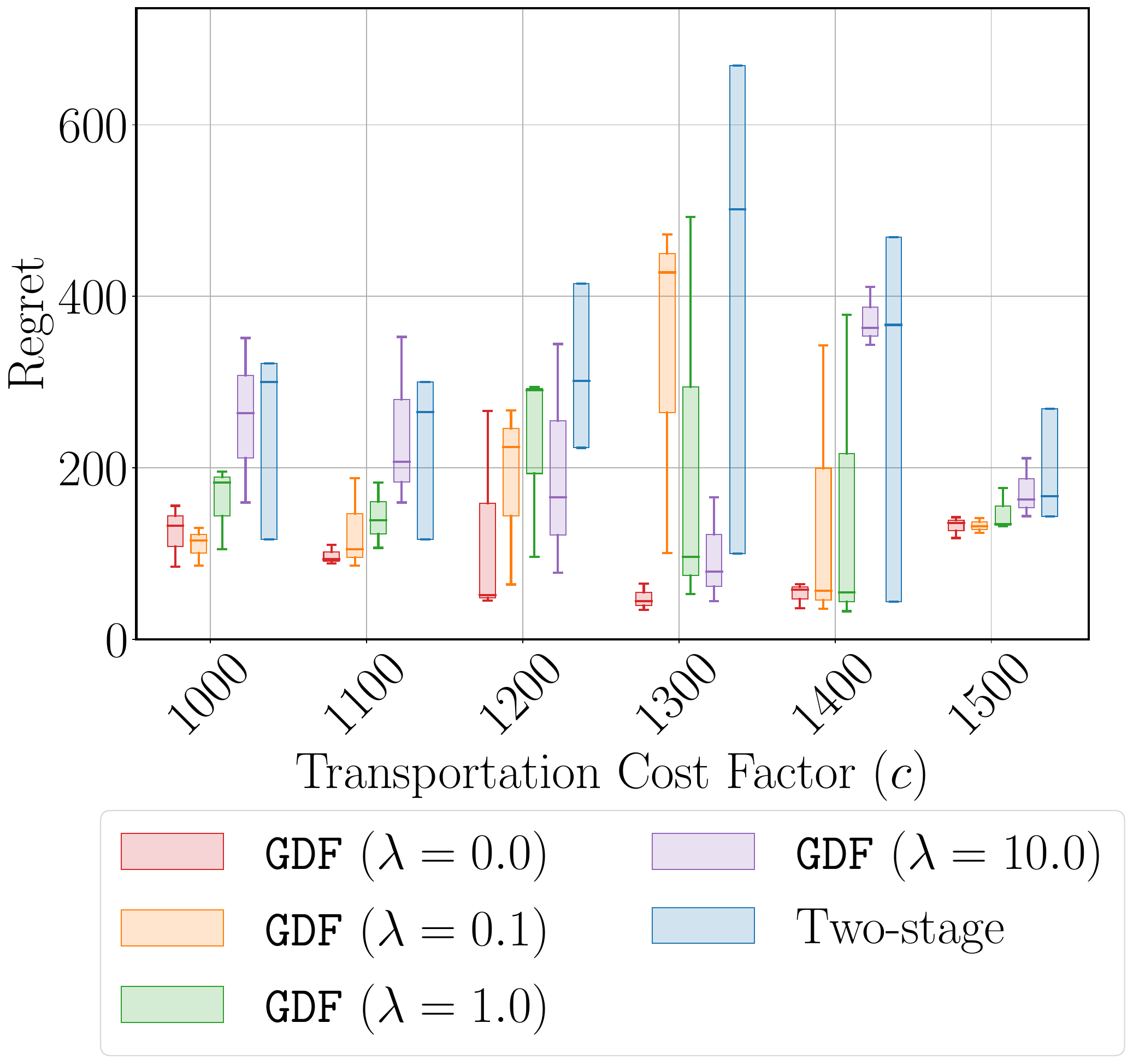}
\caption{Regret performance of the \texttt{GDF} method with varying $\lambda$ values across different transportation cost factors in synthetic data for mobile generator deployment problem, benchmarked against the Two-stage method.}
\label{fig:Total_Regret_vs_Transportation_lambda_mse}
\vspace{-.1in}
\end{figure}

\subsection{Implementation Details of Ablation Study on Neural ODE}
To ensure reproducibility and fair comparison across architectures, we summarize the model configurations and training settings used in the ablation study presented in Section~V.
The RNN model consists of two recurrent layers with 64 hidden units each, \texttt{tanh} activation, and a dropout rate of 0.2, followed by a fully connected output layer that predicts the outage trajectory over time.
The LSTM model includes two stacked LSTM layers with 64 hidden units per layer, also using a dropout rate of 0.2, and a final fully connected output layer mapping the last hidden state to predicted outages.
The Neural ODE is implemented in two variants: (i) a compact SIR-based model only for short-period prediction in ablation study with only two trainable coefficients, failure transmission rate and restoration rate, and (ii) a more complex version where $\phi(\boldsymbol{z}_k; \theta_U)$ and $\phi(\boldsymbol{z}_k; \theta_R)$ are parameterized by separate two-layer neural networks that take exogenous weather and demographic features as input.
Each network comprises a 32-unit ReLU multilayer perceptron (MLP) followed by a linear output head and Sigmoid function that enforces positive-valued rates.
The ODE dynamics are integrated using an adaptive RK4 solver implemented with \texttt{torchdiffeq}.

All RNN/LSTMs are optimized with Adam at a learning rate of $10^{-3}$ for 500 epochs.
The compact Neural ODE is trained for 500 epochs with a learning rate of $10^{-2}$, while the weather-aware Neural ODE is trained with Adam at a learning rate of $5\times10^{-4}$ for 1,000 epochs.

\subsection{Detailed Description on Online Algorithms for Mobile Generator Deployment Problem}
We include pesudocode for the proposed online baseline allocation methods in Alg.~\ref{alg:sms}.

\begin{algorithm}[H]
\caption{Observe-then-optimize algorithm for mobile generator deployment problem}
\label{alg:sms}
\textbf{Input:} Observations or forecasts $Y_{t,i}$, parameters $(\tau,\,N_g,\,\gamma)$, warehouse stock $s_w$, etc.\\
\textbf{Output:} Shipping decisions $\{x^{\text{to}}_{t,i}, x^{\text{back}}_{t,i}\}$ for each time $t$ and city $i$.
\begin{algorithmic}[1]
\FOR{$t = 1$ to $T$}
    \STATE \textit{---\;update warehouse and city stocks from previous shipments\;---}
    \FOR{each city $i$}
        \STATE Compute demand shortfall $d_{t,i} \leftarrow \max(0,\ \lceil Y_{t,i} / N_H\rceil - q_{t,i})$.
        \STATE $x^{\text{to}}_{t,i} \leftarrow \min(s_w(t),\, d_{t,i})$ \quad \textit{// send enough to cover shortfall}
        \STATE $x^{\text{back}}_{t,i} \leftarrow 0$ \quad  \textit{//  no return shipments}
    \ENDFOR
\ENDFOR
\end{algorithmic}
\end{algorithm}

\subsection{Additional Results for the Mobile Generator Deployment Problem}

This section presents additional results and visualizations for the mobile generator deployment problem.

As demonstrated in the ablation study, when $\lambda$ is large, the MSE dominates model training, reducing the advantage of \texttt{GDF} over MSE-trained models in decision quality. For more detailed ablation results in the mobile generator deployment problem, see Fig.~\ref{fig:Total_Regret_vs_Transportation_lambda_mse}, which shows that as $\lambda$ increases, the \texttt{GDF} results become similar to those of the Two-stage method, resulting in larger regret and higher variance.

Further more, Table~\ref{table:Generator Distribution Problem_synthetic} summarizes the out-of-sample performance for the generator deployment problem on synthetic data across three transportation cost factors (100, 500, and 1000). As the transportation cost increases, the improvement in decision quality for \texttt{GDF} compared to the Two-stage methods becomes more apparent. This highlights the importance of scheduling and proactive actions when transportation costs are high, demonstrating the clear advantage of \texttt{GDF}.

\begin{figure}[!t]
\centering
\includegraphics[width=\linewidth]{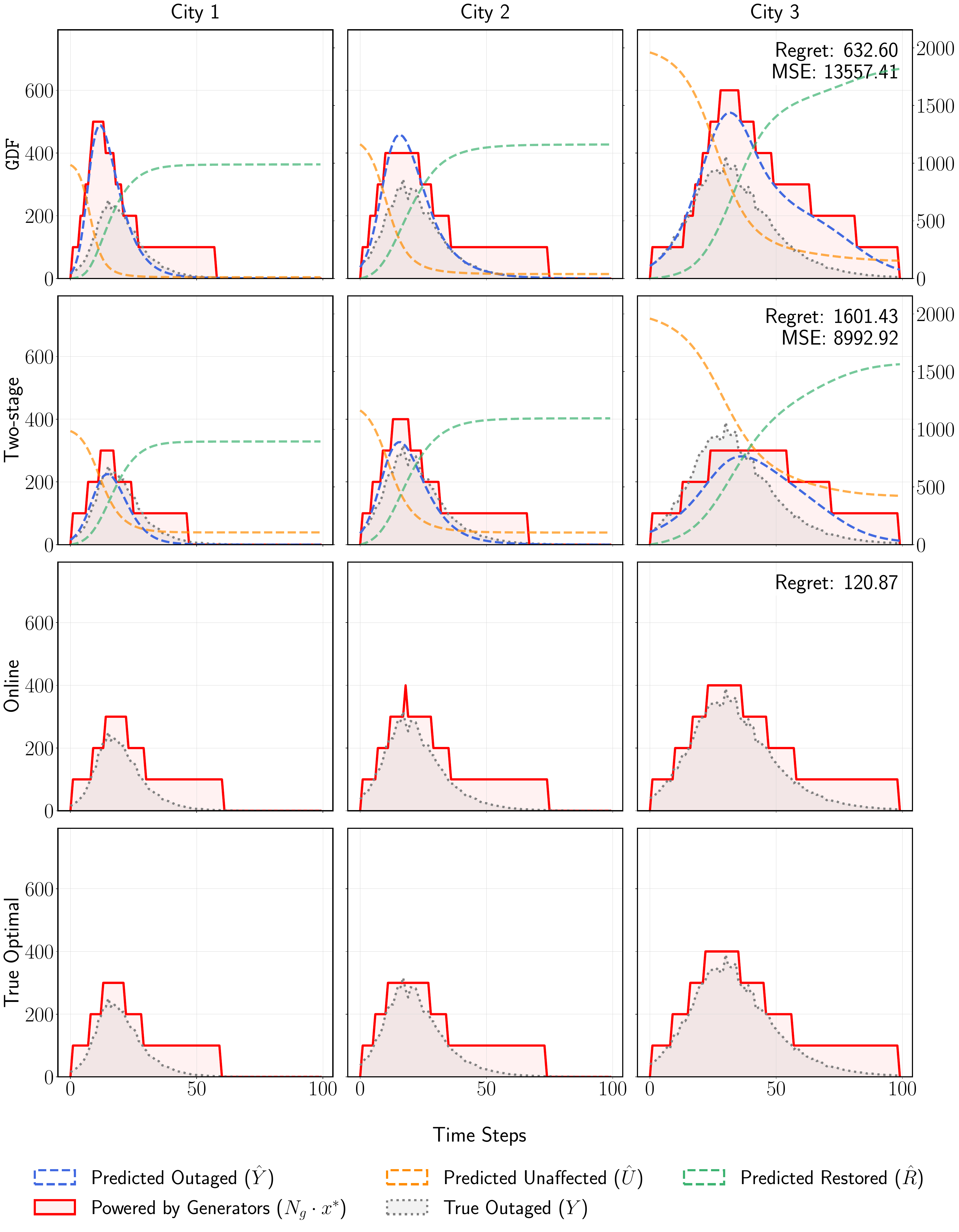}
\caption{A synthetic instance of the mobile generator deployment problem for a system with three cities and five generators ($Q_w = 10$). The $y$-axis shows the number of households experiencing outages over time. In this example, the transportation cost is set to $c = 10$, the customer interruption cost to $\tau = 1$, and the operational cost to $\gamma = 2$. The online method operates with a one-day observation lag. Travel time $\delta_t = 0$ is neglected in this case.}
\label{fig:SIR_plot_4rows_seed0_tc10_gamma2.0_G10_tau1_Ng100.0_lambda0}
\vspace{-.1in}
\end{figure}

\begin{figure}[ht]
\centering
\includegraphics[width=\linewidth]{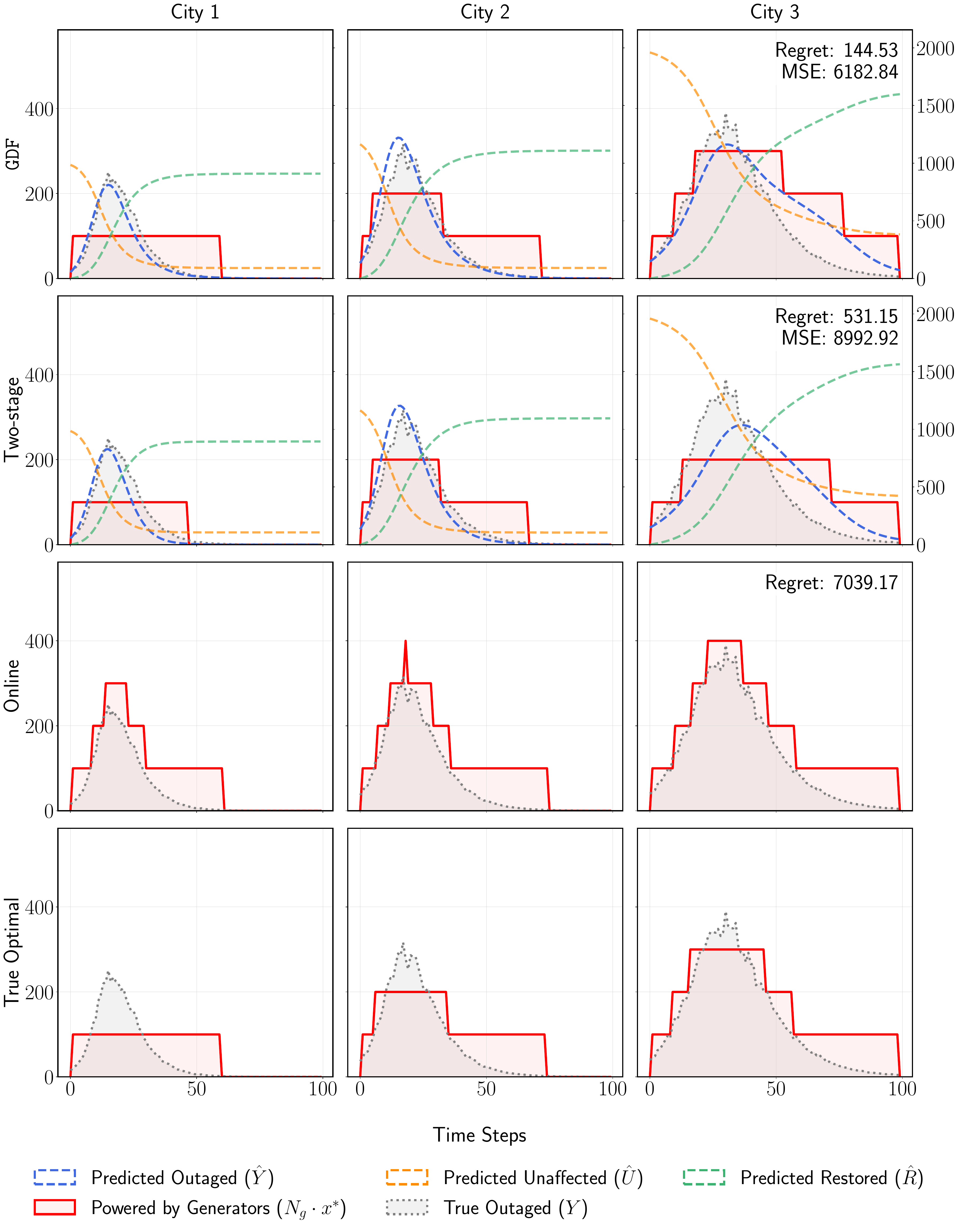}
\caption{
A synthetic instance of the mobile generator deployment problem for a system with three cities and five generators ($Q_w = 10$). The $y$-axis shows the number of households experiencing outages over time. In this example, the transportation cost is set to $c = 1000$, the customer interruption cost to $\tau = 1$, and the operational cost to $\gamma = 2$. The online method operates with a one-day observation lag. Travel time $\delta_t = 0$ is neglected in this case.
}
\label{fig:SIR_plot_4rows_seed0_tc1000_gamma2.0_G10_tau1_Ng100.0_lambda0.1}
\vspace{-.1in}
\end{figure}

We also provide additional visualizations of the deployment schemes under varying conditions. Comparing  Fig.~\ref{fig:SIR_plot_4rows_seed0_tc10_gamma2.0_G10_tau1_Ng100.0_lambda0} and Fig.~\ref{fig:SIR_plot_4rows_seed0_tc1000_gamma2.0_G10_tau1_Ng100.0_lambda0.1}, we observe that when travel costs are low, the online strategy closely approximates the optimal strategy, resulting in small regret. This is because low travel costs allow the online method to frequently move generators based on previous day data at minimal expense, leading to near-optimal regret, whereas the Two-stage and \texttt{GDF} methods rely more on predictions, and the associated noise can diminish the benefits of prediction or proactive allocation under these conditions.

\begin{figure}[!]
\centering
\includegraphics[width=\linewidth]{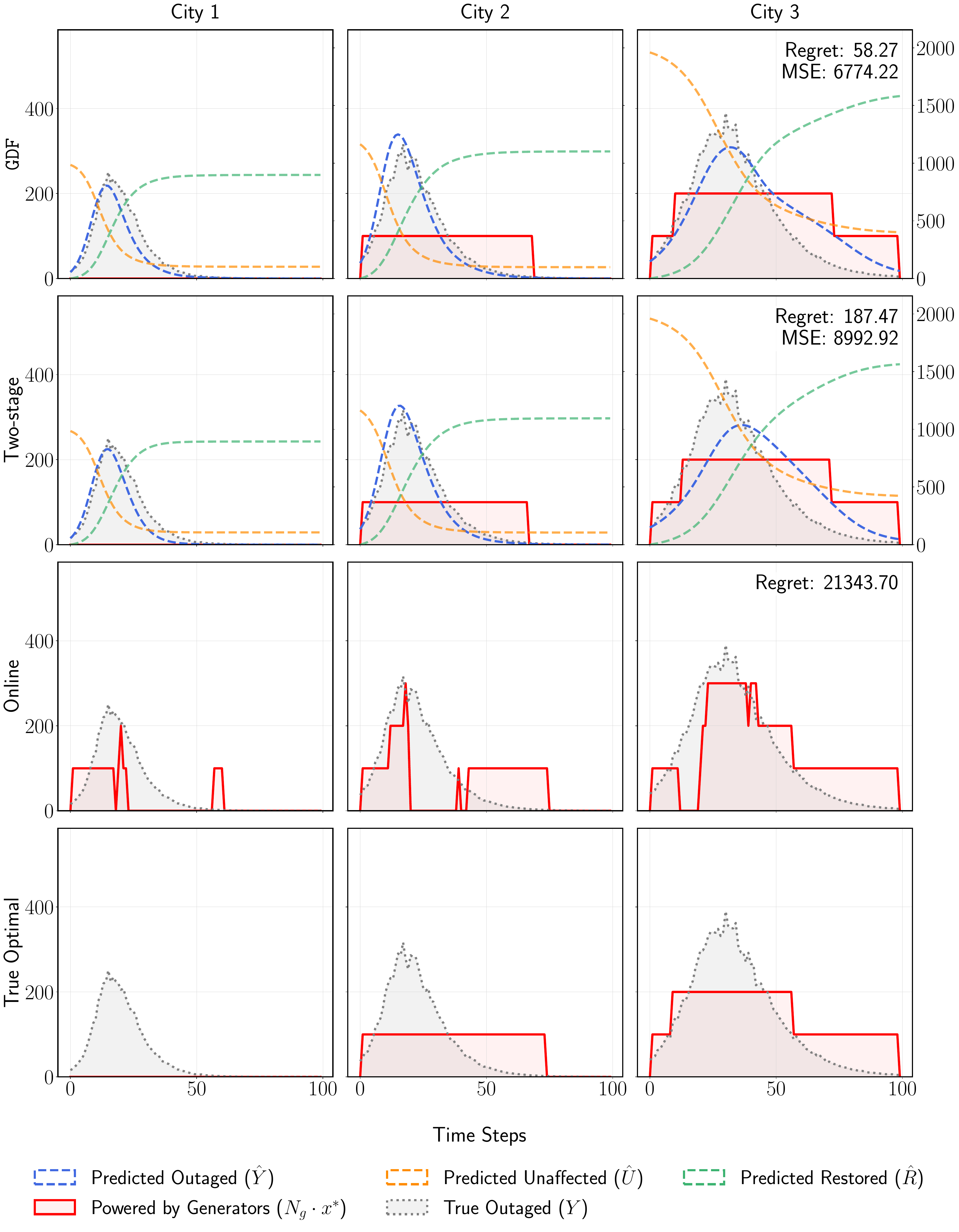}
\caption{
A synthetic instance of the mobile generator deployment problem for a system with three cities and five generators ($Q_w = 3$). The $y$-axis shows the number of households experiencing outages over time. In this example, the transportation cost is set to $c = 1000$, the customer interruption cost to $\tau = 1$, and the operational cost to $\gamma = 2$. The online method operates with a one-day observation lag. Travel time $\delta_t = 0$ is neglected in this case.
}
\label{fig:SIR_plot_4rows_seed0_tc1000_gamma2.0_G3_tau1_Ng100.0_lambda0.1}
\vspace{-.1in}
\end{figure}

In contrast, comparing Fig.\ref{fig:SIR_plot_4rows_seed0_tc1000_gamma2.0_G3_tau1_Ng100.0_lambda0.1} and Fig.\ref{fig:SIR_plot_4rows_seed0_tc1000_gamma2.0_G10_tau1_Ng100.0_lambda0.1}, we find that limited resources degrade the online strategy’s performance. Without proactive planning, fewer generators must be relocated more frequently with an online method, incurring higher transportation costs and overall regret compared to \texttt{GDF} and Two-stage approaches.

\begin{table*}[!t]
\centering
\caption{Out-of-Sample Performance for Generator Distribution problem with Synthetic Data. Results are averaged over 3 repeated
experiments with standard error (SE) in the brackets.}
\resizebox{\linewidth}{!}{%
\begin{tabular}{llllllllll}
\toprule
\multirow{2}{*}{{Model}} &
\multicolumn{3}{c}{Transportation Cost = 100} &
\multicolumn{3}{c}{Transportation Cost = 500} &
\multicolumn{3}{c}{Transportation Cost = 1000} \\
\cmidrule(lr){2-4} \cmidrule(lr){5-7} \cmidrule(lr){8-10}
& {MSE} & {Cost} & {Regret} & {MSE} & {Cost} & {Regret} & {MSE} & {Cost} & {Regret} \\
\midrule
Ground Truth & / & 6515.6 (74.4) & 0 & / & 11271 & 0 & / & 16271 & 0 \\
Online (lag = 1) & / & 8163.7 (293.4) & 1648.1 (358.6) & / & 19630.4 (1514.8) & 8358.7 (1576.8) & / & 33963.7 (3042.2) & 17692.1 (3104.4) \\
Online (lag = 3) & / & 8978.6 (277.6) & 2462.9 (313.8) & / & 22045.2 (1493.3) & 10773.6 (1522.9) & / & 38378.6 (3020.2) & 22106.9 (3050.4) \\
Online (lag = 5) & / & 9618.9 (57.3)  & 4939.9 (111.1) & / & 22152.2 (498.5) & 16497.9 (557.8) & / & 37818.9 (1074.7) & 32164.5 (1135.0) \\
Two-stage       & \textbf{9336.0 (975.6)} & 6671.5 (4328.7) & 155.8 (33.3)  & 9336.0 (975.6) & 11465.8 (94.6) & 194.2 (107.3) & 9336.0 (975.6) & 16517.8 (133.9) & 246.1 (112.7) \\
\texttt{GDF} & 9709.6 (3055.6) & 6668.7 (4365.9) & \textbf{153.0 (19.8)} & \textbf{8671.2 (1553.7)} & 11428.1 (87.1) & \textbf{156.4 (44.2)} & \textbf{6981.1 (1874.0)} & 16395.9 (89.0) & \textbf{124.2 (36.3)} \\
\bottomrule
\end{tabular}
}
\label{table:Generator Distribution Problem_synthetic}
\vspace{-.1in}
\end{table*}


\subsection{Additional Results for Power Line Undergrounding}
\begin{figure}[!t]
\centering
\includegraphics[width=\columnwidth]{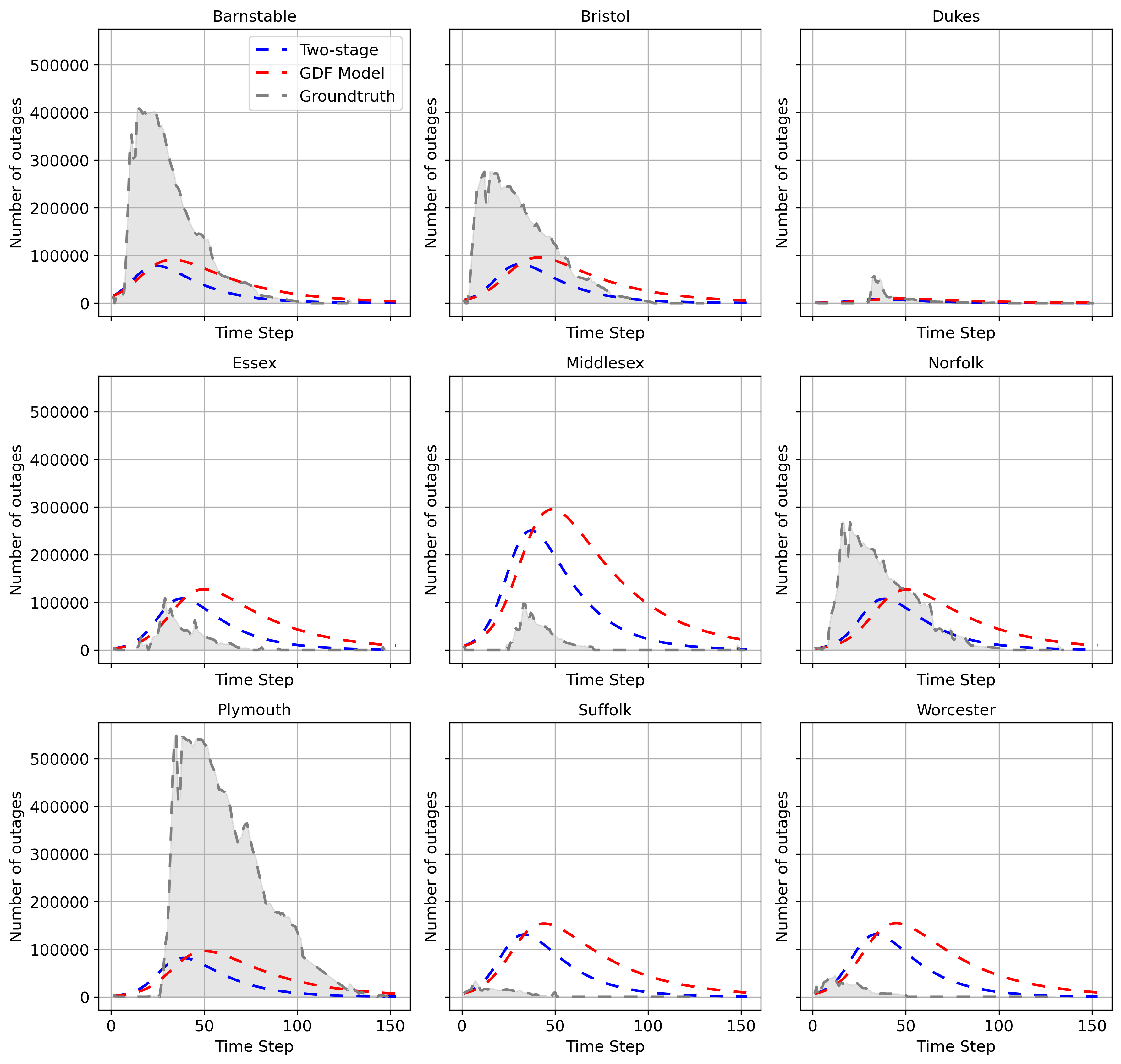}
\caption{\texttt{GDF} is overestimating outage in certain counties to prioritize resource allocation at the cost of MSE. 
}
\label{fig:hardenining_real_ma}
\vspace{-.1in}
\end{figure}

Fig.~\ref{fig:hardenining_real_ma} shows the predicted outage trajectories for all Massachusetts counties using the \texttt{GDF} model compared with groundtruth and Two-stage. Notably, the model exaggerates outages for selected counties—a deliberate strategy to prioritize resource allocation. This controlled overestimation, while slightly increasing MSE. Overall, it effectively reduces SAIDI and regret compared to the Two-stage baseline, demonstrating that decision-focused training can enhance overall scheduling performance.

\subsection{Training-time comparison across models}
\begin{figure}[h!]
    \centering
    \includegraphics[width=0.5\textwidth]{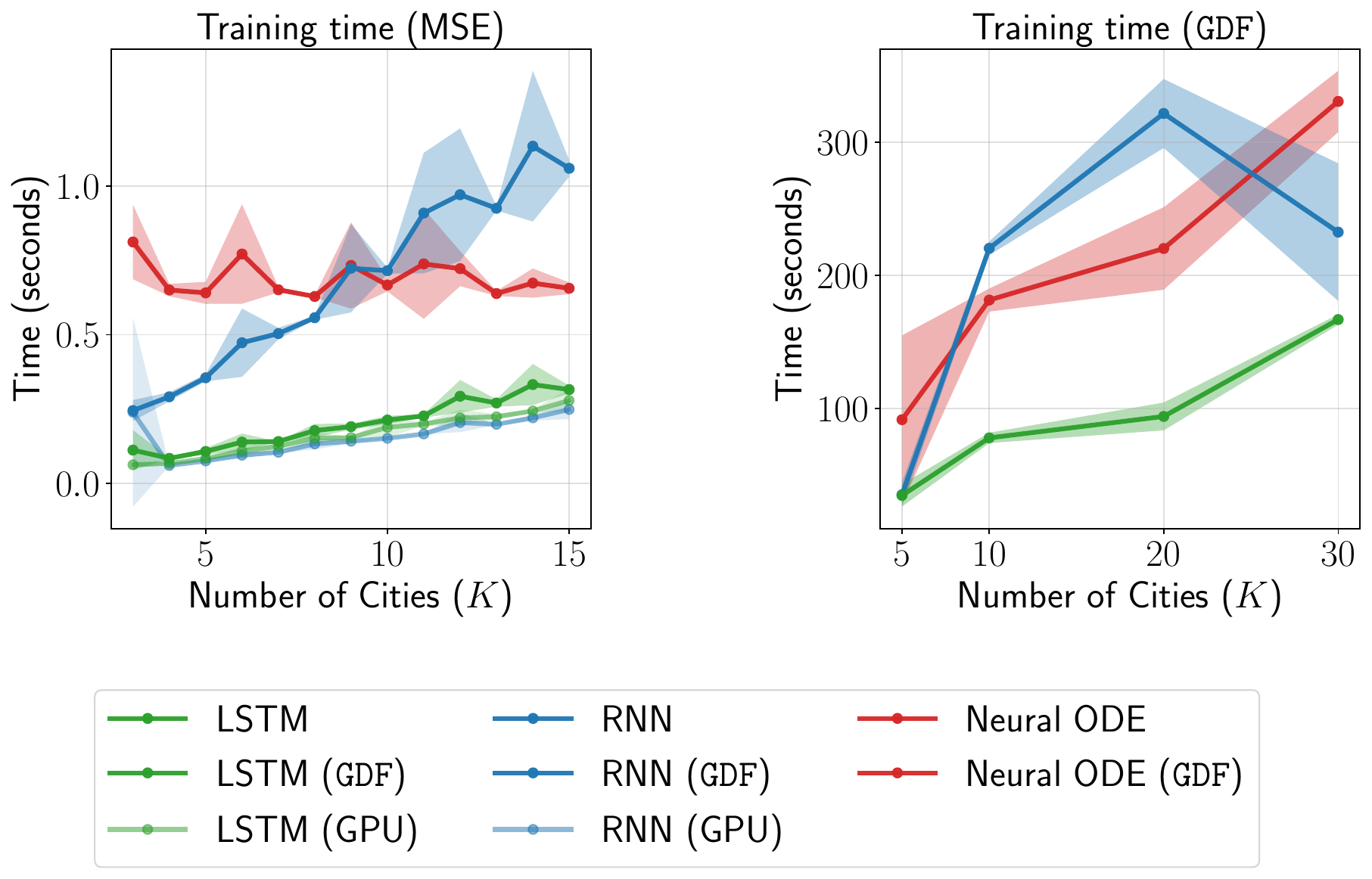} 
    \caption{Time taken by Nerual ODE, RNN, and LSTM over different number of cities: left panel shows time taken for 10 epochs of MSE training; right panel shows time taken for 10 epochs of \texttt{GDF} training. The model sizes are aligned with our setup for Massachusetts and Indianapolis case studies. For GPU experiments, we used an NVIDIA T4 on Google Colab with 16 GB GPU memory and 25 GB system RAM.}
    \label{fig:scalibility}
\end{figure}

\paragraph{Parameter complexity}
Let $d_z$ denote the weather-feature dimension, $H$ the hidden width, $L$ the number of hidden layers, and $K$ the number of service areas.  
The Neural ODE model uses two shared rate networks, yielding parameter complexity
\[
\Theta(2 (K \cdot d_z H + L H^2)),
\]
which is independent of $M$ when rates are fully shared.  
In contrast, training an RNN or LSTM separately for each city scales as
\[
\text{RNN: } \Theta\!\big(K(L d_z H + L H^2)\big), 
\quad
\text{LSTM: } \Theta\!\big(K \cdot 4L(d_z H + H^2)\big),
\]
which grows linearly with $K$ due to per-city duplication.  
Thus, as $K$ increases, the RNN and LSTM incur more significant longer training times, while Neural ODE remains efficient by sharing hidden layers across all areas.
\paragraph{Results}
Figure~\ref{fig:scalibility} compares wall-clock training times for different models in both MSE and \texttt{GDF} stages.
The left panel confirms that the Neural ODE scales gracefully with $K$, consistent with its $\Theta(K\,d_z H + L H^2)$ complexity.
By contrast, per-city RNN and LSTM show rapidly increasing costs consistent with their $\Theta(K\,\cdot)$ parameter growth, even with GPU acceleration.
These results indicate that our Neural ODE structure offers a more scalable and efficient framework than training separate time-series models per city.

The right panel of Figure~\ref{fig:scalibility} shows the growth of training times under \texttt{GDF}.
In practice, with MSE pretraining and \texttt{GDF} finetuning, the number of \texttt{GDF} epochs is typically one-third to one-tenth of the MSE epochs.

\subsection{Extended Literature Review for DFL and Differentiable Optimization}
\noindent\emph{Decision-Focused Learning}.
Decision-focused learning (DFL) has emerged as a powerful framework for integrating predictive models with downstream optimization tasks. Unlike traditional two-stage approaches, which first train standalone prediction models and then use their predictions as input parameters to optimal decision models, DFL aligns the prediction model's training loss with the objective function of the downstream optimization. This concept is enabled in gradient descent training by backpropagating gradients through the solution to an optimization problem. When the optimization is a differentiable function of its parameters, this can be implemented via implicit differentiation of optimality conditions such as KKT conditions \cite{amos2017optnet, gould2021deep} or fixed-point conditions \cite{kotary2023folded, wilder2019end}.  When the optimization is nondifferentiable, it can instead be implemented by means of various approximation techniques \cite{kotary2021end, mandi_decision-focused_2024}.  
Unlike traditional two-stage approaches, which first train standalone prediction models and then use their predictions as inputs for decision-making, DFL directly embeds the optimization problem within the learning process. This allows the learning model to focus on the variables that matter most for the final decision \cite{mandi_decision-focused_2024}.

Elmachtoub and Grigas \cite{elmachtoub_smart_2022} first proposed the \textit{Smart Predict-and-Optimize} (SPO) framework, which introduced a novel method for formulating optimization problems in the prediction process. SPO essentially bridges the gap between predictive modeling and optimization by constructing a decision-driven loss function that reflects the downstream task. However, the SPO framework only addresses linear optimization problems and does not extend well to more complex combinatorial tasks. 

The most-studied class of nondifferentiable optimization problems in decision-focused learning (DFL) involves linear programs (LPs). Notably, the Smart Predict-and-Optimize (SPO) framework by Elmachtoub and Grigas \cite{elmachtoub_smart_2022} introduced a convex surrogate upper bound to approximate subgradients for minimizing the suboptimality of LP solutions based on predicted cost coefficients. 
The most-studied class of nondifferentiable optimizations are linear programs (LPs). Elmachtoub and Grigas \cite{elmachtoub_smart_2022} proposed the Smart Predict-and-Optimize (SPO) framework for minimizing the suboptimality of solutions to a linear program  as a function of its predicted cost coefficients.
Despite this function being inherently non-differentiable, a convex surrogate upper-bound is used to derive informative subgradients. Wilder et al. \cite{wilder2019melding} propose to smooth linear programs by augmenting their objectives with small quadratic terms  \cite{amos2017optnet} and differentiating the resulting KKT conditions.
A method of smoothing LP's by noise perturbations was proposed in \cite{berthet2020learning}.
Differentiation through combinatorial problems, such as mixed-integer programs (MIPs), is generally performed by adapting the approaches proposed for LP's, either directly or on their LP relaxations. For example, Mandi et al. \cite{mandi2020smart} demonstrated the effectiveness of the SPO  method in  predicting cost coefficients to MIPs. Vlastelica et al. \cite{vlastelica2019differentiation} demonstrated their method directly on MIPs, and Wilder et al. \cite{wilder2019melding} evaluated their approach on LP relaxations of MIPs.


Building on these works, we extend DFL to spatio-temporal decision-making for power grid resilience management. Our approach employs quadratic relaxations to enable gradient backpropagation through MIPs \cite{wilder2019melding}, thereby integrating a spatio-temporal ODE model for power outage forecasting directly into the optimization process. Additionally, we introduce a Global Decision-Focused Framework that combine prediciton error with decision losses across geophysical units, improving grid resilience against extreme natural events and bridging the gap between localized predictions and system-wide decisions.


\vspace{.1in}
\noindent\emph{Differentiable Optimization}.
Differentiable optimization (DO) techniques have demonstrated significant potential in integrating predictive models with optimization problems. By enabling the computation of gradients through optimization processes, DO facilitates the seamless incorporation of complex system objectives into machine learning models, thereby enhancing decision-making capabilities \cite{cvx}.
Recent extensions of DO methods have tackled challenges beyond standard optimization tasks. For example, distributionally robust optimization (DRO) problems have been addressed using differentiable frameworks to handle prediction tasks under worst-case scenarios. For instance, \cite{zhu2022distributionally, chen2025uncertainty} employed DO-based techniques to improve uncertainty quantification and robust learning, effectively addressing data scarcity and enhancing resilience modeling.

Beyond predictive modeling, DO has advanced solutions in combinatorial and nonlinear optimization. Techniques such as implicit differentiation of KKT conditions \cite{amos2017optnet} and fixed-point conditions \cite{kotary2023folded} address differentiable constraints, while approximation methods, including noise perturbation \cite{berthet2020learning} and smoothing techniques \cite{vlastelica2019differentiation}, enable gradient computation for nondifferentiable tasks.

These advancements underscore DO’s pivotal role in bridging predictive modeling and optimization, especially where decision quality critically affects system resilience. In this work, DO is employed to align spatio-temporal outage predictions with grid optimization objectives, enabling robust strategies for generator deployment and power line undergrounding. 


\end{document}